\documentclass[lettersize,journal]{IEEEtran}
\usepackage{amsmath,amsfonts}
\usepackage{algorithmic}
\usepackage{algorithm}
\usepackage{array}
\usepackage{makecell}
\usepackage[caption=false,font=normalsize,labelfont=sf,textfont=sf]{subfig}
\usepackage{textcomp}
\usepackage{stfloats}
\usepackage{url}
\usepackage{amssymb}
\usepackage{graphicx} 
\usepackage{colortbl}
\usepackage[table]{xcolor}
\usepackage{booktabs}
\usepackage{arydshln}
\usepackage{multirow}

\usepackage{verbatim}
\usepackage{graphicx}
\usepackage{cite}
\usepackage{hyperref}
\usepackage{orcidlink} 
\hypersetup{hidelinks} 

\begin{document}

\title{JTCSE: Joint Tensor-Modulus Constraints and Cross-Attention for Unsupervised Contrastive Learning of Sentence Embeddings}

\author{Tianyu Zong$^{\orcidlink{0009-0006-1468-1750}}$, Hongzhu Yi$^{\orcidlink{0009-0003-3485-6886}}$, Bingkang Shi$^{\orcidlink{0009-0009-5389-139X}}$, Yuanxiang Wang$^{\orcidlink{0009-0003-9810-2717}}$, 

Jungang Xu$^{\orcidlink{0000-0002-3994-1401}}$, ~\IEEEmembership{Member, ~IEEE}

\thanks{
Tianyu Zong, Hongzhu Yi, Yuanxiang Wang, and Jungang Xu are with the School of Computer Science and Technology, University of Chinese Academy of Sciences, Cloud Computing and Intelligent Information Processing Laboratory(CCIP), Beijing, 101408, China. E-mail: \{zongtianyu20, yihongzhu23, wangyuanxiang19\}@mails.ucas.ac.cn, xujg@ucas.ac.cn.

Bingkang Shi is with the Institute of Information Engineering, Chinese Academy of Sciences, Beijing, 100085, China. E-mail: shibingkang20@mails.ucas.ac.cn.


}
}

\markboth{Journal of \LaTeX\ Class Files,~Vol.~14, No.~8, August~2021}%
{Shell \MakeLowercase{\textit{et al.}}: A Sample Article Using IEEEtran.cls for IEEE Journals}


\maketitle

\begin{abstract}

Unsupervised contrastive learning has become a hot research topic in natural language processing. Existing works usually aim at constraining the orientation distribution of the representations of positive and negative samples in the high-dimensional semantic space in contrastive learning, but the semantic representation tensor possesses both modulus and orientation features, and the existing works ignore the modulus feature of the representations and cause insufficient contrastive learning. 
Therefore, we first propose a training objective that is designed to impose modulus constraints on the semantic representation tensor, to strengthen the alignment between positive samples in contrastive learning.
Then, the BERT-like model suffers from the phenomenon of sinking attention, leading to a lack of attention to CLS tokens that aggregate semantic information. In response, we propose a cross-attention structure among the twin-tower ensemble models to enhance the model's attention to CLS token and optimize the quality of CLS Pooling.
Combining the above two motivations, we propose a new \textbf{J}oint \textbf{T}ensor representation modulus constraint and \textbf{C}ross-attention unsupervised contrastive learning \textbf{S}entence \textbf{E}mbedding representation framework JTCSE, which we evaluate in seven semantic text similarity computation tasks, and the experimental results show that JTCSE's twin-tower ensemble model and single-tower distillation model outperform the other baselines and become the current SOTA.
In addition, we have conducted an extensive zero-shot downstream task evaluation, which shows that JTCSE outperforms other baselines overall on more than 130 tasks.

We open source the code and checkpoints for this work as follows: \url{https://github.com/tianyuzong/JTCSE}.

\end{abstract}


\begin{figure}
    \centering
    \includegraphics[width=1\linewidth]{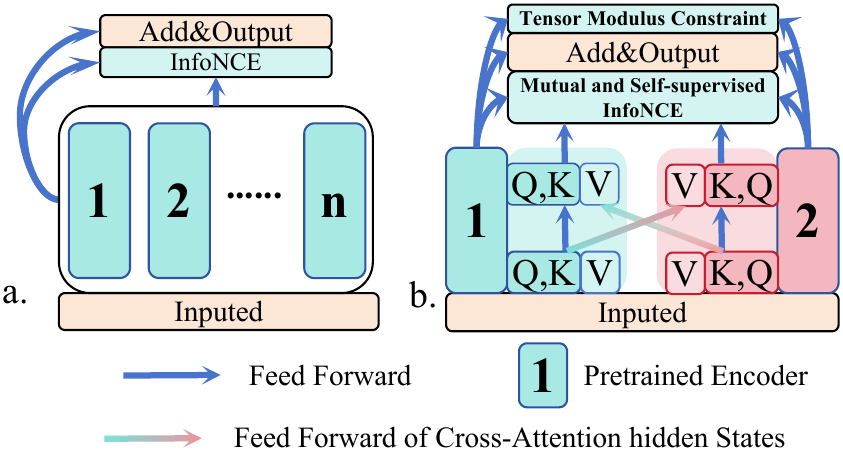}
    \caption{Subfigure \textbf{a}. represents the traditional ensemble modeling approach (EDFSE\cite{zong}), which naively trains multiple sub-encoders separately and then sums the outputs. This approach causes a large inference overhead.
    Subfigure \textbf{b.} represents the optimized ensemble learning framework JTCSE proposed in this work. It incorporates semantic representation tensor modulus constraints and joint modeling of cross-attention between sub-encoders. This framework contains only two sub-encoders. It significantly reduces inference overhead while improving the quality of sentence embeddings relative to \textbf{a.}}
    \label{intro}
    \vspace{-1em}
\end{figure}

\section{Introduction}

The study of unsupervised sentence embedding representations has been a hot topic in natural language processing, which aims to map natural language sentences into high-dimensional tensor representations that can be applied to a wide range of downstream tasks.
Some years ago, with the advent of pre-trained language models (PLMs) BERT\cite{BERT} and RoBERTa\cite{roberta}, much work has been done based on these two PLMs, e.g., Sentence-BERT\cite{sbert}, ConSERT\cite{consert}, and SimCSE\cite{gao2021simcse}.
SimCSE applies InfoNCE's\cite{infonce} idea of contrastive learning\cite{simclr} by generating positive samples through the Dropout method of the BERT-like model at training time and uniformly distributing unlabeled soft-negative samples.
With the appearance of SimCSE, many works are based on the idea of unsupervised contrastive learning in SimCSE and InfoNCE. For example, ESimCSE\cite{esimcse} augments the positive sample in SimCSE by constructing proximity words to replace individual words in the original sample. In addition, ESimCSE introduces the idea of momentum queueing in MoCo\cite{moco} to expand the scope of contrastive learning.
DiffCSE\cite{diffcse} learns the differences between original sentences and forged sentences by generating forged samples through ELECTRA\cite{electra} and the Replaced Token Detectio task to improve the quality of sentence tensor representations. ArcCSE\cite{arccse} generates multiple positive samples by masking the original sentences multiple times and constructing the positive sample triples to model entailment between sentence pairs in the triples. 
InfoCSE\cite{infocse} directly adds several EncoderLayers of Transformers\cite{attention} as auxiliary networks to the exterior of the BERT-like model and provides MLM constraints on the output of the auxiliary networks, while InfoNCE self-supervised constraints are applied to the ontology of the BERT-like model. SNCSE\cite{SNCSE} is inspired by PromptBERT\cite{PromptBERT} and uses the Prompt technique and different templates to generate enhanced positive and negative samples to improve the quality of positive and negative sample pairs in SimCSE.
EDFSE\cite{zong} first applies the data augmentation method of Round Trip Translation (RTT), which translates the original English dataset into different languages and then into English through a translation system. It then applies the idea of ensemble learning\cite{Ensemble} to train multiple BERT-like pre-trained models with different RTT datasets and the method of SimCSE, respectively. Finally, these BERT-like models are integrated into a large ensemble model to achieve model performance improvement.
RankCSE\cite{rankcse} also applies the same idea of ensemble learning, with the difference that it uses the existing checkpoint SimCSE-base/large as the teacher and ranks the similarity of the current sample to other soft-negative samples by multiple teachers and students. The multiple teachers constrain the students' performance through KL scatter loss, while the unsupervised InfoNCE still constrains the students.

However, there are some common problems with the above works, and to clarify the significance and direction of this work, we have organized the motivations as follows:

\subsubsection{\textbf{Existing works neglect the modulus feature of sentence embedding representations}}
Starting from SimCSE, the common point of the above baselines is that they all involve unsupervised contrastive learning as the primary training method and employ InfoNCE as the central loss for model training, fine-tuning a BERT-like model. However, these works invariably ignore a critical issue.
Since each EncoderLayer of the BERT-like model contains two LayerNorm layers for normalizing the sample features, the high-dimensional features of the text are mapped onto a hypersphere due to the presence of LayerNorms. This causes the text tensor representation to lose the modulus-length features and retain only the orientation features.
Meanwhile, InfoNCE only constrains the alignment of feature representations between positive samples by cosine similarity and unlabeled soft samples are distributed to the whole hypersphere, which makes the modulus feature of the text be further ignored; therefore, there exists a situation in all SimCSE-type models, i.e., ignore the features of two mutually positive samples are very different from each other in terms of modulus although their orientations are approximately the same in the high-dimensional space while we believe that the orientation and mode length of features that are positive samples of each other should be approximately the same, it is necessary to propose a loss function for the tensor modulus constraint.

\subsubsection{\textbf{More attention is needed for the CLS token}}
There is a existing work\cite{attensink} introduces the possibility that generative language models may suffer from attentional sinking, and we observe that the same phenomenon exists for BERT-like embedding models.
To the best of our knowledge, existing baseline models generally adopt the CLS pooling strategy, which utilizes the hidden state of CLS tokens to represent the semantic information of the whole sentence. However, by observing the attention score distribution, we notice that almost all baseline models suffer from an “attention sink” phenomenon: the models disproportionately focus their attention on end-of-sequence tokens at the last encoding layer. We believe this attention allocation bias leads to insufficient attention to the CLS token used to gather global information, making it difficult for the CLS token to effectively capture global semantic information and lead to lower-quality sentence embedding.

The phenomenon of attention sinking is widespread in BERT-like models\footnote{We report this finding in detail in the Fig. \ref{attention_sink}.}, which may be related to BERT pre-training. All the related work is fine-tuned based on BERT-like models, which makes it difficult to directly address the phenomenon. However, we further observe that the performance of the BERT-like model on semantic textual similarity computation tasks is positively correlated with the 2-paradigm ratio of CLS to the hidden state of other tokens in the self-attention module of the EncoderLayers.
Specifically, the larger the ratio of the 2-parameter of the tensor of the hidden state corresponding to CLS to the 2-parameter of the matrix of the hidden states of the other tokens (defined as CLS energy weights), the better the model performs. 
Therefore, we can enrich the semantic information of CLS pooling by increasing the CLS energy weights.

Noting that cross-attention\cite{crossatten} has been widely employed in multimodal learning\cite{vilbert,visualbert,bridgetower,managertower}, we consider that in textual information understanding, the attention mechanism performs inter-model cross-computation to enable the CLS token to fuse the differentiated features of the twin encoder. 
Therefore, we design a cross-attention in the proposed model to enhance the CLS energy weights, enabling the CLS pooling to aggregate better sentence semantic information and alleviating the drawbacks of attention sinking.

\subsubsection{\textbf{Large inference overheads and non-autonomous training remain challenges}}
Since both EDFSE\cite{zong} and RankCSE\cite{rankcse} adopt the ensemble learning training method, i.e., constructing a multi-tower (or twin-tower) model that unifies the feature distribution of each encoder output. However, since the size of the ensemble model proposed by EDFSE is equivalent to 6 SimCSE-BERT-bases, which imposes a considerable inference overhead, and RankCSE relies on the same type of checkpoints to do knowledge distillation, which is not an autonomous training method, the above problems are yet to be solved.



Based on the above discussion, we conclude that there are three main motivations: The first is that the modulus of high-dimensional feature representations with mutually positive samples need to be constrained; the second is that cross-attention needs to be introduced to increase the quality of CLS pooling, which in turn improves the model's performance on downstream tasks, and the third is that traditional ensemble learning methods should be optimized to reduce the inference overhead and improve the quality of the output tensor. Therefore, in this paper, we propose a joint semantic tensor modulus constraint and cross-attention in ensemble model for unsupervised contrastive learning of sentence embedding, JTCSE.

According to the first motivation, we first propose an intuitive training objective, i.e., the two semantic representation tensors of positive samples should have similar modulus and orientations; in other words, the two feature representations of positive samples should be close to each other in terms of their distributional positions in the high-dimensional space; according to the second motivation, inspired by the methods of visual-language to construct an ensemble model, we design the twin-tower model to achieve feature fusion by constructing cross-attention between towers and to promote feature sharing and information complementarity between towers, as shown in Fig. \ref{intro}, which boosts the CLS energy weights in the model and optimize the performance of CLS pooling in downstream tasks.
Incorporating the solutions to the first two motivations, we obtain an ensemble model of the twin-tower trained autonomously, which solves the third motivation by both compressing the inference overhead and avoiding the discussion of non-autonomous training.

Following the existing works, we evaluate seven semantic text similarity (STS) tasks (\cite{sts12}, \cite{13}, \cite{14}, \cite{15}, \cite{16}, \cite{stsb}, \cite{sickr}) and achieve SOTA results in the baseline of all open-source checkpoints to the best of our knowledge. Compared to the EDFSE-BERT-base, our proposed JTCSE-BERT-base has an inference overhead close to one-third of it but outperforms it on 7 STS tasks, proving our proposed framework's effectiveness.
To have a fairer comparison, we propose two approaches. The first one concerns the non-autonomously trained model RankCSE. We perform ensemble learning on the other baselines and re-compare them on the 7 STS tasks, and the results show that the JTCSE is better. Second, we compress the twin-tower JTCSE into a single-tower model employing knowledge distillation with the same parameter scales as the other baselines. On the 7 STS tasks, we get the distillation model that still performs the optimal.
We report significant experimental results showing that the proposed model's performance gain does not depend on the setting of random seeds.
In addition, we have conducted a broader zero-shot evaluation based on the MTEB\footnote{\url{https://pypi.org/project/mteb/1.28.6/}} framework, including more than 130 tasks such as text retrieval, text classification, text re-ranking, bi-textmining, multilingual text semantic similarity, etc. The results show that our proposed JTCSE and the derived models are generally ahead of the current open-source checkpoints.
In addition, we report the performance gains from tensor modulus constrained objective and cross-attention, respectively; we visualize the attentional sink phenomenon for the BERT-like model and also report the trend line of the near-positive correlation between CLS energy weights and STS task performance. In the discussion section, we present a detailed analysis of the motivation and methodological soundness of our proposed tensor modulus constraints. In addition to the significant experimental results, we report the inference overhead for different baselines and visualize the experimental results of alignment and uniformity for unsupervised embedding representation models.

We summarize the main contributions as follows:
\begin{itemize}
\item{To the best of our knowledge, we are the first to propose a modulus-constrained training objective targeting unsupervised contrastive learning, and the proposed training objective is proved to be effective through extensive comparative experiments and ablation experiments.}
\item{In order to enhance the BERT-like model's attention to CLS tokens, we introduce cross-attention in the twin-tower ensemble model, which is jointly modeled by multiple spatial mappings to enhance the energy weight of CLS pooling, and hence optimize the model's performance on multi-tasks.}

\item{Combining the tensor modulus constraints and the cross-attention mechanism, our proposed twin-tower ensemble model effectively reduces the inference overhead of the traditional multi-tower ensemble model EDFSE and performs better on the semantic textual similarity computation task.} 
\item{We have conducted extensive evaluations. Firstly, JTCSE performs best in seven semantic text similarity tasks, and our proposed JTCSE and its derived models perform best in the currently open-source checkpointing baseline in a variety of 0-shot evaluations for downstream tasks in natural language processing. We conduct a detailed ablation analysis to gain insight into the strengths of the proposed model and open source the entire code and checkpoints of this work.}

\item{To promote research progress in related areas, we have open-sourced the code and saved checkpoints for this work.} 

\end{itemize}

This work is an extension of the existing work TNCSE: Tensor's Norm Constraints for Unsupervised Contrastive Learning of Sentence Embeddings\cite{zong2025tncse}, which has been accepted by AAAI25 for Oral presentation. This work's main update compared to previous work is the addition of a cross-attention structure to the twin encoder to enhance the BERT-like model's attention to the CLS token. This strengthens the CLS token's ability to capture the global semantic information of the sentence and optimizes the model's CLS pooling performance in downstream tasks.
We report the main updates to this work relative to the predecessor work as follows:
\begin{itemize}
    \item \textbf{New Movation:} We have noticed that unsupervised sentence embedding models of BERT-like models usually suffer from attention sinking; they pay more attention to the end punctuation or SEP token of the input sequence in the last coding layer, and lack of attention to CLS token, coupled with the fact that all of these baselines use CLS pooling, we believe should enhance the model's attention to CLS token attention to improve the quality of CLS pooling.

    \item  \textbf{New Method:} Directly optimizing the attention weight matrix may destroy the pre-training information of BERT-like models. For this reason, we propose the concept of CLS energy weights to enhance the model's attention to the CLS token by boosting the CLS energy weights. Based on existing works, we find that different EncoderLayers of BERT-like models focus on different features of the input sequence. Intuitively, different EncoderLayers in different models may also capture different semantic features of the same sentence. Therefore, we introduce a cross-attention mechanism for feature fusion between models. This approach enriches the semantic information aggregated by the CLS token, which enhances the CLS energy weight and thus optimizes the performance of CLS Pooling in downstream tasks.

    \item  \textbf{New Experimental Results:} Relative to the previous work TNCSE, the model JTCSE retrained in this work has improved its performance on 7 STS tasks; in order to evaluate JTCSE's generalization ability more comprehensively, we have conducted an extensive 0-shot evaluation of downstream tasks of natural language processing, 
    our evaluation shows JTCSE achieves average performance gains across more than 130 downstream tasks compared to existing baselines, reaching new SOTA results.

    \item  \textbf{Other Updates:} We enrich the insight of the tensor modulus constrained training objective design by decomposing it into two sub-objectives and discussing their significance separately. For the more extensive evaluations, we add the English part of the three multilingual STS tasks and enrich the seven STS tasks evaluated by existing work into ten. The results show that JTCSE and the derived models still generally perform best.
    
\end{itemize}
In the subsequent sections, this paper systematically summarizes the representative work on unsupervised sentence embedding models in the related work section and overviews the typical applications of the cross-attention mechanism in multimodal information fusion; in the method section, we derive the tensor modulus constraint training objective in detail based on the motivation of solving the problem that the InfoNCE loss function neglects the positive samples' modulus alignment, and meanwhile targeting to alleviate the attention sinking and enhance the CLS Pooling information density, an innovative cross-attention structure is proposed, and finally, the model architecture and loss function design are presented in full;
the Experiments section details the training data and evaluation tasks and reports the performance of the model on more than 130 tasks; the Ablation study discusses the impact of the components and the reasonableness of the loss function design; the Discussion section concerns the reasonableness of the tensor-module constrained training objectives' design and detailed motivations for cross-attention design.

\begin{figure}
    \centering
    \includegraphics[width=0.75\linewidth]{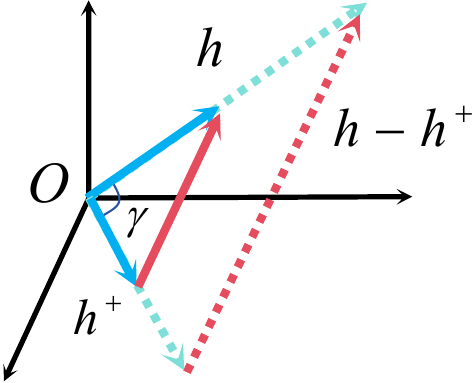}
    \caption{This figure represents the distribution of the positions of a pair of positive sample semantic representation tensors $h$ and $h^{+}$ in three-dimensional space and the vectors $h-h^{+}$ for which they are subtracted.
    According to the principle of similar triangles, when the angle $\gamma$ is specific, the larger the modulus of $h$ or $h^{+}$, the larger the modulus of $h-h^{+}$ will be, and the greater the value of being constrained.}
    \label{pami2}
    \vspace{-1em}
\end{figure}

\begin{figure}
    \centering
    \includegraphics[width=0.75\linewidth]{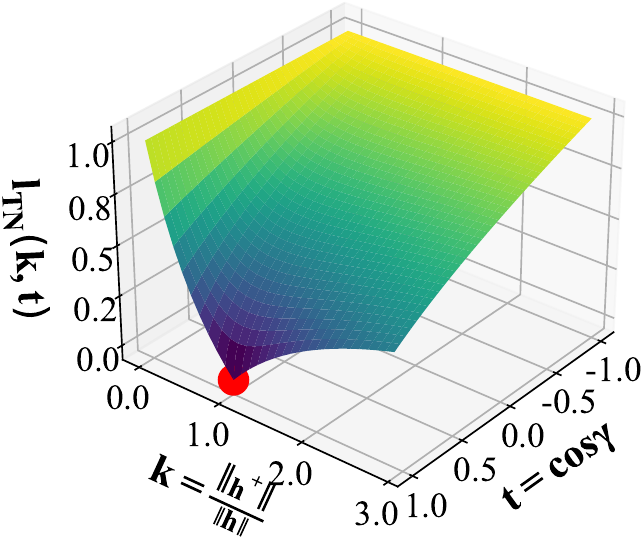}
    \caption{This figure illustrates the binary loss function $L_{TMC}$, with respect to the range of values of the two independent variables $t$ and $k$ over part of its domain of definition.}
    \label{pami3}
    \vspace{-1em}
\end{figure}

\section{Related Work}

\subsection{Unsupervised Sentence Embedding Approach}

InfoNCE\cite{infonce} (Noise Contrastive Estimation Loss) is a widely used loss function for self-supervised learning, mainly for feature representation learning. The method usually uses cosine similarity to compare the similarity between positive and negative samples and optimize the model parameters. In unsupervised sentence embedding training, many studies have combined InfoNCE loss with pre-trained language models (e.g., BERT\cite{BERT}, RoBERTa\cite{roberta}), which has pushed the progress of contrastive learning.
For example, both SimCSE\cite{gao2021simcse} and ConSERT\cite{consert} utilize the idea of dropout to generate positive samples, and both use cosine similarity as the only metric to distinguish between positive and negative samples. ConSERT further introduces various data augmentation strategies (e.g., random deletion, random insertion, etc.\cite{eda}) to enrich positive samples. Subsequent studies have continuously optimized the training methods for unsupervised sentence embedding based on SimCSE.
For example, ESimCSE\cite{esimcse} combines near-synonym data augmentation and MoCo's\cite{moco} momentum queuing mechanism to improve the quality of SimCSE's representations; DiffCSE\cite{diffcse} combines masked language modeling by introducing an additional discriminator ELECTRA\cite{electra} and further optimizes the model performance by employing InfoNCE; ArcCSE\cite{arccse} refers to the positive-negative-sample triple construct proposed by SentenceBERT\cite{sbert} and fine-tuned based on SentenceBERT; SNCSE\cite{SNCSE} employs a comparison learning strategy with soft negative samples combined with bi-directional marginal loss; InfoCS\cite{infocse} additionally introduces an additional network for mask language modeling; EDFSE\cite{zong} employs the Round-Trip Translation data augmentation strategy to train multiple encoders to construct a large-scale integration model; and the current SOTA method RankCSE\cite{rankcse} utilizes dual-teacher ensemble learning with distillation techniques to train encoders.
The common point of the above studies is that they all introduce the InfoNCE loss. However, they only rely on the cosine similarity between embeddings when using InfoNCE for similarity metrics, ignoring the critical factor of the modulus of the embedding tensor.

To this end, we address this motivation by proposing an unsupervised training objective incorporating embedding representation modulus constraints to improve further the model's ability to detect positive and negative sample discrimination.

\subsection{Application of Cross-Attention}

Cross-attention has been widely used in the field of multimodal embedding alignment.Visual-BERT\cite{visualbert} encodes image and text tokens into a multimodal sequence that is fed into an Encoder-Only model for joint multimodal modeling; 
VILBERT\cite{vilbert} uses a dual-stream structure to process visual and linguistic information separately before feature fusion, LXMERT\cite{LXMERT} processes the inputs of the two modalities separately and introduces a text mask, and introduces a cross-modal encoder in addition to a self-encoder for each modality, and ALIGN\cite{align} uses contrastive learning to align the image and text into a shared embedding space.
CLIP\cite{clip} does not directly apply cross-attention and further optimizes performance on the visual-verbal task through a simple twin-tower structure and intuitive multimodal contrastative learning; BridgeTower\cite{bridgetower} and ManagerTower\cite{managertower} build on CLIP by introducing an additional cross-attention network between the twin towers and applying an early feature fusion strategy, and on the Visual-Linguistic Question and Answer task outperforms CLIP.

Based on the research trend of multimodal learning, we have found that the better the performance of the twin-tower ensemble model in a visual-linguistic alignment task, the more parameters need to be introduced for cross-attention-based feature fusion to achieve complementary modeling among features.

However, to our knowledge, there is no ensemble model based on cross-attention for feature sharing with a twin-tower structure in unsupervised sentence embedding. Thus, it is necessary to propose it in this field. We start from the perspective of compressing the training overhead as much as possible, similar to CLIP, by keeping only the basic encoder without introducing other training parameters, and based on cross-attention to achieve feature complementarity and enhance the representation quality of CLS Pooling.



\section{Methods}

In this section, we first introduce the proposed training objective for semantic representation tensor constraints, then introduce the cross-attention structure design, and finally describe the proposed JTCSE's overall structure and the loss function's design.

\subsection{Modulus length constraints of semantic tensor representations}

Existing methods for training sentence embedding representations of unsupervised contrastive learning usually evaluate the correlation of a pair of samples by their cosine similarity. During training, the model is trained on the distribution of tensor representations between positive and negative samples by the constraints of the InfoNCE loss function, which constrains the cosine similarity between them to be as large as possible for pairs of positive samples and as uniformly distributed as possible for unlabeled soft negative samples.
From the perspective of positive samples, InfoNCE requires the orientation of positive sample tensor pairs to be aligned; however, from the mathematical point of view, a tensor has the features of "magnitude" and "orientation," while InfoNCE only constrains the tensor's "orientation" but ignores the "magnitude."
SimCSE applies InfoNCE by passing a sample through a BERT-like model to get the corresponding tensor, and due to the presence of Dropout in the model, a sample can produce two similar features, which are positive samples to each other, and the features of the other samples in a mini-batch are as soft-negative samples for unsupervised training. 
SimCSE and its derivatives use the “orientation” of the tensor as the only metric to judge the similarity between positive and negative samples, which lacks attention to the “magnitude” of the tensor. Therefore, we use the modulus of the tensor, i.e., the 2-parameter, to represent the “magnitude” of the tensor, and from an intuitive geometric perspective, we propose a constraint objective $L_{TMC}$ for the modulus of the tensor between pairs of positive samples, in order to strengthen the model to judge the features of positive and negative samples that do not have apparent differences in orientation as shown in Eq. \ref{ltmc}.
\begin{equation}
    L_{TMC}(h,h^{+} ) = \frac{\left \| h-h^{+}  \right \| }{\left \| h \right \|+\left \| h^{+}  \right \|  } .
    \label{ltmc}
\end{equation}
In $L_{TMC}$, $h$ and $h^{+}$ denote the features of a pair of positive samples, respectively, and $\left \| h  \right \|$, $\left \| h^{+}  \right \|$  denotes the modulus of the tensor, which is the 2-parameter. We first construct the tensor representation space of a pair of positive samples and difference vectors according to Fig. \ref{pami2}, and we expect the model to be trained with the angle $\cos \gamma $ and the modulus of the difference vectors of both as small as possible.

In addition, intuitively, the larger the modulus of $\left \| h  \right \|$ and $\left \| h^{+}  \right \|$ is, the more pronounced the modulus of their difference vectors are and the more valuable the constraints are when the angles $\gamma $ are equal, so we establish two sub-objectives. The first is that the modulus of the difference vectors of the positive sample pairs should be as small as possible, and the second is that the modulus of each of the positive sample pairs should be as large as possible. Therefore, we construct Eq. \ref{ltmc} with the sum of the modulus of $\left \| h  \right \|$ and $\left \| h^{+}  \right \|$ as the denominator and the modulus of the difference vectors of both pairs as the numerator. During training, both sub-objectives are optimized simultaneously; from a quantitative point of view, $L_{TMC}$ has no measure and can be combined with other loss functions.

To more rigorously justify $L_{TMC}$, we make a simple transformation. Firstly, according to Fig. \ref{pami2} and the cosine theorem, $\left \| h  \right \|$, $\left \| h^{+}  \right \|$, and $h-h^{+} $ can construct a closed triangle, and $\left \| h-h^{+}  \right \| $ can be rewritten as Eq. \ref{h-h+}:
\begin{equation}
    \left \| h-h^{+}  \right \| = \sqrt{\left \| h \right \|^{2}+\left \| h^{+}   \right \|^{2} -2\cdot \left \| h \right \| \cdot \left \| h^{+}  \right \| \cdot \cos \gamma    } ,
    \label{h-h+}
\end{equation}
$L_{TMC}$ can be rewritten as Eq. \ref{ltmcp}: 
\begin{equation}
    L_{TMC}  = \frac{\sqrt{\left \| h \right \|^{2}+\left \| h^{+}   \right \|^{2} -2\cdot \left \| h \right \| \cdot \left \| h^{+}  \right \| \cdot \cos \gamma    } }{\left \| h \right \| + \left \| h^{+}  \right \| } .
    \label{ltmcp}
\end{equation}
Since the tensor modulus of the samples are all larger than zero, there exists Eq. \ref{kh+}: 
\begin{equation}
    \left \| h^{+}  \right \| =k\cdot \left \| h \right \| , k\in \left ( 0,+\infty  \right ) .
    \label{kh+}
\end{equation}
Moreover, since $\cos \gamma$ takes the value of $\left [ -1,1 \right ] $, let $t=\cos \gamma $, $t\in \left [ -1,1 \right ] $, so Eq. \ref{ltmcp} can be further rewritten as Eq. \ref{ltnkt}:
\begin{equation}
    L_{TMC}\left ( k,t \right )  = \frac{\sqrt{1+k^{2} -2\cdot k\cdot t} }{1+k}. 
    \label{ltnkt}
\end{equation}
More intuitively, we visualize the binary function $L_{TMC}$ as shown in Fig. \ref{pami3}.

It can be seen from Fig. \ref{pami3} that when $L_{TMC}$ obtains the minimum value, $k = 1$ and $t = 1$, i.e., $\left \| h \right \| = \left \| h^{+} \right \|$ and $\cos \gamma = 1$ , which is in accordance with our intended training objective, meaning that the tensors of positive samples of each other should have similar modulus and should have similar orientations.
Thus, we demonstrate that the proposed $L_{TMC}$ is consistent with our first propose motivation.

\subsection{Designing for Cross-Attention}

We observe that BERT-like models suffer from attention sinking, where the model disproportionately focuses on the SEP token or punctuation at the end of the sentence rather than the CLS token in deeper encoderlayers, which is detrimental to unsupervised sentence embedding models relying on CLS pooling.
To quantify this, we define the CLS energy weight $E_{CLS}$
defined as Eq. \ref{CLS's energy weight}:
\begin{equation}
    E_{CLS} = \frac{\left \| h_{cls}  \right \|_{2}  }{\left \| H_{-}  \right \| _{F} } ,
\label{CLS's energy weight}    
\end{equation}
which represents the ratio of the CLS token’s 2-norm to the Frobenius norm of other tokens’ hidden states in the context tensor. A higher $E_{CLS}$ indicates richer semantic aggregation by the CLS token and correlates with better sentence embedding performance. 
To enhance $E_{CLS}$, we introduce a cross-attention mechanism within a twin-encoder architecture. This mechanism enables interaction between encoders by using one encoder’s attention weights to weigh the other’s Value tensor, which enriches the CLS representation with complementary semantic information without disrupting the original attention distribution or BERT’s pre-trained knowledge, thus improving the global representation quality of the CLS token.

We describe briefly the motivation for introducing cross-attention through the above\footnote{In the Discussion section, we will elaborate on the motivations and details of cross-attention.}, and the process of computing cross-attention will be described in detail in the next step.

In order not to cause additional training overhead, we do not introduce other naive network weights and only consider the attention weights inside the EncoderLayer within both sub-encoders. We define the network weights involved in computing the cross-attention in each sub-encoder as $X_{N}^{i}$, where $X$ denotes the attention network, $X\in \left \{ Q,K,V \right \}$, $N$ denotes the source sub-encoder, $N\in \left \{ \mathbf{I},\mathbf{II} \right \}$ , and $i$ denotes the ordinal number of the EncoderLayer that $X$ comes from, $i \in \left [ 1, 12\right ]$\footnote{This work is oriented to the BERT-base model with 12 EncoderLayers.}.

\begin{figure*}
    \centering
    \includegraphics[width=1\linewidth]{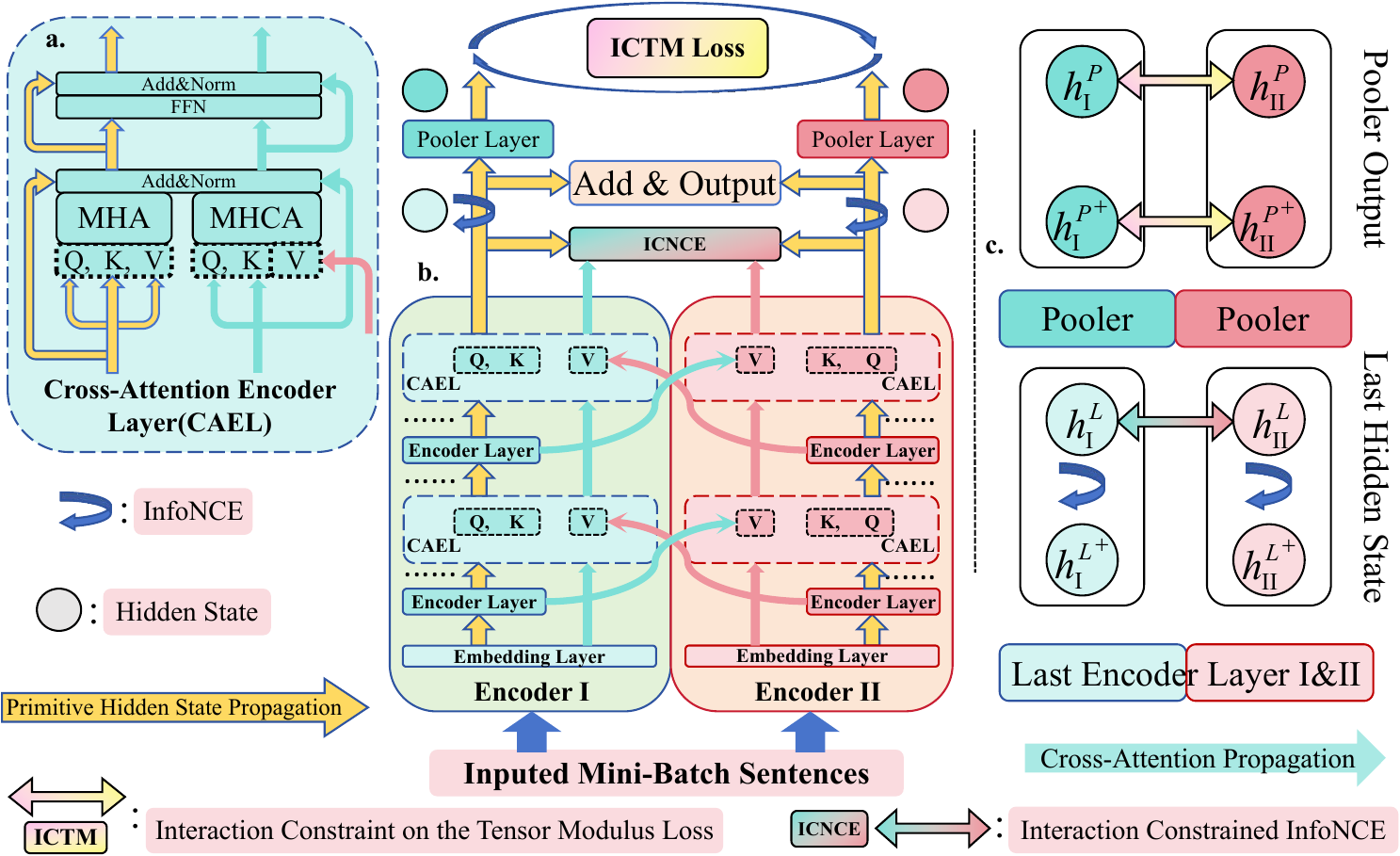}
    \caption{This figure shows the structure of the proposed unsupervised sentence embedding representation framework, JTCSE, which consists of two main parts: the semantic representation tensor modulus constraints and the joint modeling of subencoder cross-attention.
    Subfigure \textbf{b.} shows the overall structure of JTCSE, which contains two subencoders, I and II. Each is a fine-tuned BERT-like model that includes an embedding layer, an encoder, and a pooler layer. Before the training, we specify the \textbf{c}ross-\textbf{a}ttention \textbf{e}ncoder \textbf{l}ayer's(CAEL) position in the encoder, the position of CAEL in both subencoders is the same.
    During training, a mini-batch is fed into the embedding layer of two sub-encoders simultaneously, and the hidden state output from each embedding layer goes into its own encoder; if CAEL is encountered, in addition to the normal forward propagation within each sub-encoder, it is also necessary to mutually pass through the attention network in each other's EncoderLayer to achieve the computation of cross-attention.
    Both the primitive last hidden state (LHS) and the cross-attention's LHS pass through the IC-InfoNCE constraints. The primitive LHS also passes through the pooler layer to get the pooler output, which in turn passes through the tensor modulus-constrained loss function.
    Subfigure \textbf{a.} represents the details of CAEL, the Query, Key, and Value weights in MHA and MHCA are identical. Subfigure \textbf{c.} represents the details of ICTM loss and IC-InfoNCE loss.}
    \label{pami4}
    \vspace{-1em}
\end{figure*}

We first specify the location of the cross-attention layer, as shown in Eq. \ref{eq6}:
\begin{equation}
    \mathbf{L}=\left \{ i|i\in \mathbf{Z} \mathrm{} ,1\le i\le 12 ,i\bmod k=0 \right \} ,
    \label{eq6}
\end{equation}
where $\mathbf{L}$ denotes the set of all locations where the cross-attention appears in the EncoderLayer of sub-model, and $k$ is a hyperparameter denoting the presence of a \textbf{c}ross-\textbf{a}ttention \textbf{E}ncoder\textbf{L}ayer (abbreviated as CAEL, shown in Fig. \ref{pami4}a) every $k$ EncoderLayers.

$HD_{\mathbf{I}}^{j}$ and $HD_{\mathbf{II}}^{j}$ denote the output of the $j$-th EncoderLayer when the hidden state is forward propagated within the two encoders. 
When $j\in \mathbf{L}$, $HD_{\mathbf{I}}^{j}$, 
and $HD_{\mathbf{II}}^{j}$ are to perform the cross-attention computation, and for Encoder $\mathbf{I}$, 
the $Q_{\mathbf{I}}^{j}$ and $K_{\mathbf{I}}^{j}$ networks first process $HD_{\mathbf{I}}^{j-1}$ and compute the self-attention matrix $SA_{\mathbf{I}}^{i}$, as shown in Eq. \ref{SAMI}:
\begin{equation}
SA_{\mathbf{I} } ^{j} =\mathbf{softmax} \left ( \frac{(Q_{\mathbf{I} }^{j-1}\times HD_{\mathbf{I} }^{j-1} ) \cdot  (K_{\mathbf{I} }^{j-1}\times HD_{\mathbf{I} }^{j-1} )^{\mathbf{T} }}{\sqrt{d} }   \right ) ,
    \label{SAMI}
\end{equation}
where $d$ denotes the hidden state dimension.
Then, the output of $V_{\mathbf{I}}^{j}$ is weighted to obtain the context tensor $CT_{\mathbf{I}}^{j}$ inside Encoder $\mathbf{I}$, as shown in Eq. \ref{CTI}:
\begin{equation}
    CT_{\mathbf{I} }^{j}=SA_{\mathbf{I} }^{j}\cdot  (V_{\mathbf{I} }^{j} \times HD_{\mathbf{I} }^{j-1})
    \label{CTI}.
\end{equation}
Meanwhile, $SA_{\mathbf{I} } ^{i}$ also weights the output of $V_{\mathbf{II}}^{j}$ in Encoder $\mathbf{II}$ to obtain the \textbf{c}ross-\textbf{a}ttention \textbf{c}ontext \textbf{t}ensor(CACT), as shown in Eq. \ref{CACTI}:
\begin{equation}
    CACT_{\mathbf{I} }^{j}=SA_{\mathbf{I} }^{j}\cdot  (V_{\mathbf{II} }^{j} \times HD_{\mathbf{II} }^{j-1})
    \label{CACTI}.
\end{equation}
The above operation is the same for Encoder $\mathbf{II}$. According to the previous analysis, $SA_{\mathbf{I} } ^{j}$ may focus on local information. In contrast, $V_{\mathbf{II} }^{j} \times HD_{\mathbf{II} }^{j-1}$ focuses on global information, so the CLS token enriched with semantic information in Encoder $\mathbf{I}$ not only extracts information from this encoder's local context but also obtains complementary information from the global context of Encoder $\mathbf{II}$. The vice versa is valid for the CLS token in Encoder $\mathbf{II}$.
This aggregation of multi-source information makes the hidden state of the CLS token more diversified, which enhances the $E_{CLS}$ and the model's focus on the CLS token.

\subsection{Model Structure Design}
In this subsection, we introduce the twin-tower ensemble model JTCSE based on tensor modulus constraints and cross-attention, as shown in Fig. \ref{pami4}b. The large-scale ensemble model EDFSE-BERT uses six SimCSE-BERT-bases fine-tuned with multi-language RTTs, which are designed to enrich the distribution of textual semantic representations with an "intrinsic rank," which in turn enhances the model's ability to discriminate between two similar sentences.
We follow the EDFSE approach but compose the ensemble model using only two sub-encoders augmented with RTT data to reduce the huge inference overhead.

\subsubsection{\textbf{Loss of Cross-Attention and Model Continuation Training}}

Each encoder contains EmbeddingLayers, EnocderLayers, and PoolerLayers.
We denote the two sub-encoders in JTCSE-BERT or JTCSE-RoBERTa as Encoder I and Encoder II, respectively.
Since each sub-encoder contains a Dropout function, a sample will be encoded twice by each of the two sub-encoders after entering JTCSE, resulting in a total of four tensor representations. We use $h_{\mathbf{I} }^{L}$, $h_{\mathbf{I} }^{L^{+} }$, $h_{\mathbf{II} }^{L} $, and $h_{\mathbf{II} }^{L^{+} }$ to represent the CLS pooling of last hidden state from two sub-encoders. Since these four features represent the same sample, they are mutually positive samples. We design the interaction-constrained InfoNCE (ICNCE) based on InfoNCE. the InfoNCE is represented as Eq. \ref{infon}.

\begin{equation}
    L_{N C E}\left(h_{i}, h_{i}^{+}\right)=-\log \frac{e^{ \frac{\operatorname{sim}\left(h_{i}, h_{i}^{+}\right)}{\tau } }}{e^{ \frac{\operatorname{sim}\left(h_{i}, h_{i}^{+}\right)}{\tau } }+\sum e^{ \frac{\operatorname{sim}\left(h_{i}, h_{i}^{+}\right)}{\tau } }} ,
    \label{infon}
\end{equation}
where $i\in \left \{ \mathbf{I} ,\mathbf{II} \right \}$ indicates from Encoder I or Encoder II. $h_{i}$ and $h_{i}^{+}$ denote the representation of the current pair of positive samples, and $h_{i}^{-}$ denotes the representation of the current mini-batch of other soft negative samples. We set the temperature coefficient $\tau = 0.05$ according to SimCSE and derived work, $sim\left ( \cdot \right )$ denotes the cosine similarity, and ICNCE is defined as Eq. \ref{icnce}.
\begin{align}
    L_{ICNCE} = R\cdot \left ( L_{NCE}\left ( h_{\mathbf{I} }^{L},h_{\mathbf{II} }^{L}  \right ) + L_{NCE}\left ( c_{\mathbf{I} }^{O},c_{\mathbf{II} }^{O}  \right ) \right ) 
    \notag
    \\
    +(1-R)\cdot \left ( L_{NCE}\left ( h_{\mathbf{II} }^{L},h_{\mathbf{I} }^{L}  \right ) + L_{NCE}\left ( c_{\mathbf{II} }^{O},c_{\mathbf{I} }^{O}  \right ) \right ) 
    \label{icnce}
\end{align}
$R\in \left \{ 0,1 \right \}$ denotes a binary random number. $c_{\mathbf{I} }^{O}$ and $c_{\mathbf{II} }^{O}$ denote the outputs of the last CAEL sourced from Encoder I and Encoder II. During the training process, we still employ InfoNCE to continue training Encoder $\mathbf{I}$ and Encoder $\mathbf{II}$.
We employ InfoNCE to further optimize the representation quality of the CLS token by maximizing the similarity between positive sample pairs while uniformizing the similarity between soft negative sample pairs, with the aim of keeping the two Encoders continuously trained. ICNCE is designed to allow the CLS token to focus more on the comprehensive semantic features of the input sequence, thus enhancing its ability to serve as a global representation.

\subsubsection{\textbf{Refinement of Tensor Modulus-Constrained Training Objective}}
We find that if the tensor modulus constraint loss is computed with the last hidden state of Encoder $\mathbf{I}$ and Encoder $\mathbf{II}$, the features are normalized due to the passage through the NormLayer of the last Encoderlayer, which makes the features normalized and loses the modulus features. However, we observe that the last layer of the network of the BERT-like model is a FFN named PoolerLayer; the last hidden state regained the modulus features after passing through the PoolerLayer. Thus, we naturally adopt the Pooleroutput as the input of the tensor modulus constraint loss\footnote{We will discuss this design motivation in more detail in the Discussion section.}.

Further, the overall loss function will be in the form of a summation of several sub-loss terms. To avoid subsequently optimizing the factors of the sub-loss terms, we add dynamic complementary coefficients to Eq. \ref{ltmc}: $- \log_{}{ sim\left ( h_{\mathbf{I} }^{L},h_{\mathbf{II} }^{L} \right )  }$. Amend to Eq. \ref{eq11}:
\begin{equation}
    L_{TMC}\left ( h_{i},h_{j}^{+}  \right ) =-\log \left ( sim \left ( h_{\mathrm {I}}^{L},{h_{\mathrm {II}}^{L}}   \right )  \right )  \frac{\left \| h_{i}^{P}-h_{j }^{P^{+} }    \right \| }{\left \| h_{i}^{P}  \right \| +\left \| h_{j }^{P^{+} }   \right \| } ,
    \label{eq11}
\end{equation}
where $i,j\in \left \{ \mathrm {I},\mathrm {II}       \right \} $ and $i\ne j$. $h_{\mathrm {I}   }^{P} $, ${h_{\mathrm {I}   }^{P} }^{+} $, $h_{\mathrm {II}   }^{P} $, and ${h_{\mathrm {II}   }^{P} }^{+} $ denote the pooler output of these hidden states, respectively.
Intuitively, when the tensor $h_{i}^{P}$ and $h_{i}^{P^{+}}$ are in similar directions but have significant differences in modulus, the coefficients $\log_{}{sim(\cdot )}$ are not significantly helpful, but the partition $\frac{\left \| h_{i}^{P}-h_{j }^{P^{+} }    \right \| }{\left \| h_{i}^{P}  \right \| +\left \| h_{j }^{P^{+} }   \right \| }$ compensates for the loss. 
The two product terms in Eq. \ref{eq11} are jointly constrained when both the modulus and direction differences between $h_{i}^{P}$ and $h_{i}^{P^{+}}$ are large.

Since we need to optimize the two sub-encoders jointly, we define an \textbf{i}nteraction \textbf{c}onstraint on the \textbf{t}ensor \textbf{m}odulus(ICTM), denoted as Eq. \ref{ICTN}.
\begin{equation}
    L_{ICTM}=L_{TN}\left ( h_{\mathrm {I}}  ,h_{\mathrm {II}    }^{+} \right ) +L_{TN}\left ( h_{\mathrm {II}}  ,h_{\mathrm {I}    }^{+} \right ) 
    \label{ICTN}.
\end{equation}
The purpose of $L_{ICTM}$ is to strengthen the alignment of the two sub-encoders to the modulus of the positive sample tensor in the high-dimensional space. 

Finally, we define the complete loss function for JTCSE as shown in Eq. \ref{allloss}:
\begin{equation}
    L=\sum_{i\in \left \{ \mathrm {I},\mathrm {II}       \right \} }L_{NCE}\left ( h_{i}^{L},{h_{i}^{L}}^{+}   \right ) +L_{ICNCE}+L_{ICTM} .
    \label{allloss}
\end{equation}

We will prove the necessity of each component of Eq. \ref{allloss} in ablation experiments.
In addition, the ablation experiment on pooling proves that the last hidden states outperform the pooler outputs. Thus, during inference in JTCSE, the two sub-encoders will encode the input samples separately and sum the two obtained last hidden states as the output of the whole model without the need for pooler layer processing.

\section{Experiments}

\subsection{Setup}
In JTCSE, we employ two sub-encoders previously fine-tuned by the RTT training set generated by the Google Translate system and the unsupervised SimCSE. For the training dataset, we choose a 1M wiki corpus\footnote{\url{https://huggingface.co/datasets/princeton-nlp/datasets-for-simcse}} and an unsupervised SICKR dataset.\footnote{When we use the official code and retrain some existing open-source work, we find that the results reproduced according to the default hyper-references of the official code are 2\%$\sim $3\% lower than the reported results, and when additional unsupervised SICKR datasets are added, the reproduced results are barely equal to the reported results, and to be fair, we add unlabeled SICKR datasets when training TNCSE and JTCSE, and reported the effects of the unlabeled SICKR dataset in the ablation experiments.} Following the previous work, we take the Spearman correlation in the STS-B\cite{stsb} validation set as the checkpoint-saving metric.
We report the training hyperparameter settings and RTT setup details in Appendix \textbf{I}.

\begin{table*}[ht!]
\begin{center}
\caption{The left table reports the results of the JTCSE and baseline evaluation on the seven STS tasks, and the right table reports the performance of the open-source representative baseline on the English subtask of the three multilingual STS tasks.
$\star $ and $\diamondsuit $ denote results derived from the original paper and \cite{gao2021simcse}, respectively. \colorbox{blue!20}{\makebox[1.5em][l]{\textbf{\textcolor{blue!20}{ss}}}} indicates a fusion with existing unsupervised checkpoint knowledge. \colorbox{green!35}{\makebox[1.5em][l]{\textbf{\textcolor{green!35}{ss}}}} indicates the best result on the main metric \textbf{Avg}. Since RankCSE\cite{rankcse} has not officially open-sourced any code or checkpoints, $\clubsuit $ denotes the result of a third-party open-source code replication. \textbf{D} denotes distillation to a single encoder. }
\label{bigresult}
\begin{minipage}{0.68\textwidth} 
\begin{center}
\setlength\tabcolsep{5pt}
\begin{tabular}{lcccccccc}
\Xhline{2pt}
\textbf{Model} & \textbf{STS12} & \textbf{STS13} & \textbf{STS14} & \textbf{STS15} & \textbf{STS16} & \textbf{STSB} & \textbf{SICKR} & \textbf{7 Avg.} \\ \Xhline{1pt}
\multicolumn{9}{c}{\textbf{BERT-base}} \\ \hline
BERT-base$\diamondsuit $       & 39.70 & 59.38 & 49.67 & 66.03 & 66.19 & 53.87 & 62.06  &  56.70 \\
BERT-whitening$\diamondsuit$   & 57.83 & 66.90 & 60.90 & 75.08 & 71.31 & 68.24 & 63.73 & 66.28 \\
IS-BERT\cite{isbert}$\diamondsuit$                  & 56.77          & 69.24          & 61.21          & 75.23          & 70.16          & 69.21          & 64.25           & 66.58 \\
SBERT-base(Sup)$\diamondsuit $ & 70.97 & 76.53 & 73.19 & 79.09 & 74.30 & 77.03 & 72.91  & 74.89\\ 
ConSERT\cite{consert}$\star$         & 64.64 & 78.49 & 69.07 & 79.72 & 75.95 & 73.97 & 67.31  & 72.74  \\
SimCSE\cite{gao2021simcse}$\star$           & 68.40 & 82.41 & 74.38 & 80.91 & 78.56 & 76.85 & 72.23  & 76.25  \\
DiffCSE\cite{diffcse}$\star$          & 72.28 & 84.43 & 76.47 & 83.90 & 80.54 & 80.59 & 71.23  & 78.49  \\
ESimCSE\cite{esimcse}$\star$          & 73.40 & 83.27 & 77.25 & 82.66 & 78.81 & 80.17 & 72.30  & 78.27  \\
ArcCSE\cite{arccse}$\star$           & 72.08 & 84.27 & 76.25 & 82.32 & 79.54 & 79.92 & 72.39  & 78.11 \\
InfoCSE\cite{infocse}$\star$          & 70.53 & 84.59 & 76.40 & \textbf{85.10} & \textbf{81.95} & {82.00} & 71.37  & 78.85\\
PromptBERT\cite{PromptBERT}$\star$                & 71.56 & 84.58 & 76.98 & 84.47 & 80.60 & 81.60 & 69.87  & 78.54 \\
PCL\cite{pcl}$\star$                & 72.84 & 83.81 & 76.52 & 83.06 & 79.32 & 80.01 & 73.38  & 78.42 \\
SNCSE\cite{SNCSE}$\star$                & 70.67 & 84.79 & 76.99 & 83.69 & 80.51 & 81.35 & \textbf{74.77}  & 78.97 \\
WhitenedCSE\cite{whencse}$\star$        &74.03 &\textbf{84.90} &76.40 &83.40 &80.23 &81.14 &71.33 &78.78 \\ 
PromCSE\cite{promcse}$\star$          & 73.03	&85.18	&76.70	&84.19	&79.69	&80.62	&70.00	&78.49 \\ \hdashline
EDFSE\cite{zong}$\star$            & 74.48          & 83.14          & 76.39          & 84.45 & 80.02          & \textbf{81.97} & 72.83           & 79.04  \\
\textbf{TNCSE}   & \textbf{75.52} & 83.91 & {77.57} & {84.97} & 80.42 & 81.72 & 72.97  &  \textbf{79.58} \\

\textbf{JTCSE}   & {74.95} & {84.21} & \textbf{77.79} & 84.75 & 80.41 & 81.88 & 73.92  & \cellcolor{green!25} \textbf{79.70}  \\ \hdashline

EDFSE D\cite{zong}$\star$    & 74.50 & 83.61          & 76.24          & 84.02          & 80.44          & 81.94          & 74.16  & 79.27 \\ 
\textbf{TNCSE D}   & \textbf{75.42}  & 84.64 & {77.62} & {84.92} & 80.50 & 81.79 & 73.52  &  79.77 
 \\ 
\textbf{JTCSE D}   & {75.01}  & {84.86} & \textbf{77.76} & 84.62 & 80.38 & \textbf{82.05} & 74.53  & \cellcolor{green!25} \textbf{79.89} \\ \hline
\cellcolor{blue!20} RankCSE\cite{rankcse}$\clubsuit  $   & 74.61  & 85.70 & 78.09 & 84.64 & 81.36 & 81.82 & 74.51  & 80.10
 \\ \hdashline
\cellcolor{blue!20} RankCSE+UC   & 73.29  & \textbf{85.90} & 78.16 & 85.90 & \textbf{82.52} & 83.13 & {73.36}  & 80.32
\\ 
\cellcolor{blue!20} \textbf{TNCSE+UC }           & \textbf{75.79}  & 85.27 & {78.67} & \textbf{85.99} & 82.01 & {83.16} & 73.01  & \textbf{80.56} 
 \\ 

\cellcolor{blue!20} \textbf{JTCSE+UC }           & {75.44}  & 85.34 & \textbf{78.75} & {85.93} & 82.00 & \textbf{83.21} & \textbf{73.52}  & \cellcolor{green!25} \textbf{80.60}  \\ \hdashline

\cellcolor{blue!20} RankCSE+UC \textbf{D}   & 72.99  & \textbf{85.72} & 77.73 & 84.93 & \textbf{81.86} & 82.43 & \textbf{74.35}  & 80.00  \\ 
\cellcolor{blue!20} \textbf{TNCSE+UC D} & \textbf{75.95}  & 85.31 & \textbf{78.50} & \textbf{85.69} & \textbf{81.86} & \textbf{83.03} & 73.89  & \cellcolor{green!25} \textbf{80.60} 
\\ 

\cellcolor{blue!20} \textbf{JTCSE+UC D} & {75.22}  & 85.46 & \textbf{78.50} & {85.50} & 81.55 & {83.02} & 74.24  & \textbf{80.50} 
 \\ 
\Xhline{1.5pt}
\multicolumn{9}{c}{\textbf{RoBERTa-base}} \\ \Xhline{1.5pt}
RoBERTa-base$\diamondsuit $    & 40.88 & 58.74 & 49.07 & 65.63 & 61.48 & 58.55 & 61.63 & 56.57 \\
RoBERTa-whitening\cite{bertwhitening}$\diamondsuit $        & 46.99          & 63.24          & 57.23          & 71.36          & 68.99          & 61.36          & 62.91           & 61.73       \\
SimCSE\cite{gao2021simcse}$\star$           & 70.16 & 81.77 & 73.24 & 81.36 & 80.65 & 80.22 & 68.56  & 76.57
 \\
DiffCSE\cite{diffcse}$\star$          & 70.05 & 83.43 & 75.49 & 82.81 & 82.12 & 82.38 & 71.19  & 78.21 
 \\
ESimCSE\cite{esimcse}$\star$          & 69.90 & 82.50 & 74.68 & 83.19 & 80.30 & 80.99 & 70.54  & 77.44
 \\
PromptBERT\cite{PromptBERT}$\star$                & 73.94 & {84.74} & {77.28} & {84.99} & 81.74 & 81.88 & 69.50  & 79.15 
\\
PCL\cite{pcl}$\star$                & 71.13 & 82.38 & 75.40 & 83.07 & 81.98 & 81.63 & 69.72  & 77.90 
\\
SNCSE\cite{SNCSE}$\star$                & 70.62 & 84.42 & 77.24 & 84.85 & 81.49 & \textbf{83.07} & 72.92  & 79.23 
\\
WhitenedCSE\cite{whencse}$\star$   &70.73 &83.77 &75.56 &81.85 &\textbf{83.25} &81.43 &70.96 &78.22 \\ 
IS-CSE\cite{iscse}$\star$           & 71.39 & 82.58 & 74.36 & 82.75 & 81.61 & 81.40 & 69.99  & 77.73 \\ \hdashline

EDFSE\cite{zong}  & 72.67 & 83.00 & 75.69 & 84.07 & 82.01 & 82.53 & 71.92  & 78.84 \\         
\textbf{TNCSE}    & 74.11 & 84.00          & 76.06          & 84.80 & 81.61          & 82.68          & 73.47           & \textbf{79.53} 
 \\  
\textbf{JTCSE}    & \textbf{74.92} & 84.22          & 77.08          & 84.69 & 81.39          & 82.60          & 74.03           & \cellcolor{green!25} \textbf{79.94} \\  \hdashline
EDFSE D\cite{zong}           & 71.04          & 81.08          & 77.04          & 83.08          & 81.96          & 82.36          & \textbf{74.54}  & 78.73  \\ 
\textbf{TNCSE D}     & {74.56} & {84.74} & 76.30 & 84.89 & 81.70 & 83.01 & 74.18  & \textbf{79.91}
  \\
\textbf{JTCSE D}     & \textbf{75.42} & \textbf{85.36} & \textbf{77.31} & \textbf{85.04} & 81.72 & 82.91 & 74.46  & \cellcolor{green!25} \textbf{80.32}  \\ \hline
\cellcolor{blue!20} RankCSE\cite{rankcse}$\clubsuit  $        & 69.09          & 81.15 & 73.62 & 81.31          & 81.43          & 81.22          & 70.08           & 76.84  \\ \hdashline
\cellcolor{blue!20} RankCSE+UC     & 74.18 & 84.06 & {77.72} & 83.26 & 79.81 & 81.25 & 72.58  & 78.98 \\
\cellcolor{blue!20} \textbf{TNCSE+UC}     & {74.52} & {85.26} & 77.63 & \textbf{85.85} & \textbf{82.62} & \textbf{83.65} & {73.35}  & \textbf{80.41} 
\\ 
\cellcolor{blue!20} \textbf{JTCSE+UC}     & \textbf{74.57} & \textbf{85.73} & \textbf{78.17} & {85.78} & {82.73} & {83.73} & \textbf{73.52}  & \cellcolor{green!25} \textbf{80.61} 
 \\ 
\hdashline
\cellcolor{blue!20} RankCSE+UC \textbf{D}     & 68.55 & 82.23 & 73.61 & 81.28 & 81.28 & 80.98 & 71.01  & 76.99 
\\
\cellcolor{blue!20} \textbf{TNCSE+UC D}     & {74.14} & {83.86} & {76.09} & {84.07} & {81.59} & {82.90} & {73.55}  & \textbf{79.46} 
 \\

\cellcolor{blue!20} \textbf{JTCSE+UC D}     & \textbf{74.92} & \textbf{85.14} & \textbf{77.07} & \textbf{84.59} & \textbf{81.71} & \textbf{83.18} & \textbf{74.50}  & \cellcolor{green!25} \textbf{80.16}\\
\Xhline{2pt}
\end{tabular}
    
\end{center}
\end{minipage}%
\hspace{0.01\textwidth} 
\begin{minipage}{0.25\textwidth} 
\begin{center}
\setlength\tabcolsep{3pt}
\begin{tabular}{cccc}
\Xhline{2pt}
\textbf{STS17} & \textbf{STS22} & \textbf{STSBM} & \textbf{3 Avg.} \\ \Xhline{1pt}
\multicolumn{4}{c}{\textbf{BERT-base}} \\ \hline
- & - & - & - \\
- & - & - & - \\
- & - & - & - \\
- & -  & -  & - \\
- & -  & -  & - \\
 83.90 & 59.74 &	82.45 &	75.36 \\
 80.15 &	61.84 &	84.56 &	75.52 \\
 85.63 & 	61.33 & 	80.15 & 	75.70 \\
 - & -  & -  & - \\
85.05 	& 55.51 &	85.49 &	75.35 \\
51.33 &	50.58 &	43.75 	& 48.55 \\
\textbf{86.32} & 	63.10 & 	83.83 	& 77.75 \\
 53.11 &	54.77 	&55.55 &	54.48 \\
 85.15 &	60.83 	&84.50 &	76.83 \\
  - & -  & -  & - \\ \hdashline
 - & -  & -  & - \\
 85.78 &	61.45 &	84.14 &	77.13 \\
85.88 &	62.79 &	85.40 & \cellcolor{green!25}	\textbf{78.02} \\ \hdashline
- & -  & -  & - \\
85.36 	& \textbf{63.79} &	85.41 &	78.19 \\
85.65 	&63.59 	& \textbf{85.55} 	& \cellcolor{green!25} \textbf{78.26} \\ \hline
 85.88 &	62.46 &	62.46 &	70.26 \\ \hdashline
86.19 &	59.24 	&86.28 &	77.24 \\
86.51 	& \textbf{62.05} 	& \textbf{86.35} 	&\textbf{78.30} \\
\textbf{86.89} &	61.87 &	\textbf{86.35} &\cellcolor{green!25}	\textbf{78.37} 
 \\ \hdashline
81.62  &  60.58 	  & 81.77 & 	74.66  \\
\textbf{86.38} 	& \textbf{63.28} &	\textbf{86.24} &\cellcolor{green!25}	\textbf{78.63} \\
86.28 	& 63.06 & 	83.01 	& 77.45  \\

\Xhline{1.5pt}
\multicolumn{4}{c}{\textbf{RoBERTa-base}} \\ \Xhline{1.5pt}
- & - & - & - \\
- & -  & -  & -  \\
81.80 &	58.23 &	84.45 &	74.83 \\
82.21 &	60.90 &	84.99 &	76.03 \\
83.15 & 	60.79 	&85.36 &	76.43 \\
74.57 	&53.65 &	70.46 &	66.23 \\
81.80 &	61.58 &	85.25   & 	76.21  \\
77.26 & 	59.26 & 	83.11 & 	73.21 \\
- & -  & -  & - \\
- & -  & -  & - \\ \hdashline
- & -  & -  & - \\
\textbf{84.05} &	62.70 &	83.03 &	76.59 \\
83.73 & \textbf{	63.78} & 	\textbf{86.33} & \cellcolor{green!25}	\textbf{77.95} \\ \hdashline
 - & -  & -  & - \\
84.01 &	64.06 &	86.25 &	78.11 \\
83.18 &	\textbf{64.69} &	\textbf{86.49} &\cellcolor{green!25}	\textbf{78.12} \\ \hline
81.29 & 58.63 & 84.11 & 74.68  \\ \hdashline
82.39 & 60.32 & 85.77 & 76.16  \\
\textbf{85.49} 	&62.00 &	86.80 &	78.09 \\
85.30 	& \textbf{62.87} 	&\textbf{86.82} &	\cellcolor{green!25} \textbf{78.33} \\ \hdashline
82.57 &	60.12 &	84.32 	&75.67 \\
\textbf{84.04} &	62.72 &	86.07 &	77.61 \\
83.79 &	\textbf{64.74} &	\textbf{86.27} &\cellcolor{green!25}	\textbf{78.27} \\
\Xhline{2pt}
\end{tabular}
\end{center}
\end{minipage}
\end{center}
\vspace{-2em}
\end{table*}

\begin{table*}[h]
\setlength\tabcolsep{3pt}
\begin{center}
\caption{This table reports the results of zero-shot testing for 45 text classifications, with the optimal results on each task \textbf{bolded} and the sub-optimal results \underline{underlined}.}
\vspace{-1em}
\begin{tabular}{lccccccccccc}
\toprule[2pt]
\textbf{Tasks}                                   & \textbf{Sim$\sim $} & \textbf{ESim$\sim $} & \textbf{Diff$\sim $} & \textbf{Info$\sim $} & \textbf{SN$\sim $} & \textbf{Whiten$\sim $} & \textbf{Rank$\sim $} & \textbf{TN$\sim $} & \textbf{JT$\sim $} & \textbf{JT$\sim $ D} & \textbf{JT$\sim $ UCD} \\ \toprule[1pt]
\textbf{AllegroReviews}                                                                                             & 24.15           & 23.72            & \textbf{25.25}   & 24.13            & 24.38          & \underline{24.94}          & 24.89            & 23.59          & 23.51          & 24.05             & 24.21               \\
\textbf{AngryTweets}                                                                                                & 42.34           & 41.37            & \underline{42.54}      & 41.68            & \textbf{44.30} & 41.29                & 42.33            & 41.51          & 40.66          & 40.93             & 41.45               \\
\textbf{\begin{tabular}[c]{@{}l@{}}ContractNLIInclusionOfVerbally\end{tabular}}     & 53.96           & \underline{64.75}      & 53.96            & 61.87            & \textbf{66.19} & 48.92                & 52.52            & 56.83          & 58.27          & 54.68             & 52.52               \\
\textbf{\begin{tabular}[c]{@{}l@{}}ContractNLIPermissibleAcquirement\end{tabular}} & 79.21           & \underline{83.15}      & 82.58            & 78.65            & 74.16          & \textbf{87.08}       & 81.46            & 82.58          & 82.58          & 82.58             & \underline{83.15}         \\
\textbf{\begin{tabular}[c]{@{}l@{}}ContractNLIPermissibleDevelopment\end{tabular}} & 78.68           & \underline{88.24}      & 85.29            & 85.29            & 79.41          & \textbf{90.44}       & 83.09            & 79.41          & 85.29          & 82.35             & 85.29               \\
\textbf{CUADAntiAssignmentLegalBench}                                                                               & \textbf{84.73}  & 82.76            & 80.89            & 82.00            & 79.10          & 80.46                & 83.11            & \underline{83.70}    & 81.66          & 82.51             & 81.14               \\
\textbf{CUADExclusivityLegalBench}                                                                                  & 66.40           & 70.47            & 63.39            & 63.78            & 64.04          & 68.50                & 62.86            & 64.04          & 70.60          & \textbf{73.10}    & \underline{72.31}         \\
\textbf{CUADNoSolicitOfCustomersLegalBench}                                                                         & \textbf{84.52}  & \textbf{84.52}   & 79.76            & 76.19            & 77.38          & \textbf{84.52}       & \textbf{84.52}   & 82.14          & \textbf{84.52} & 83.33             & 83.33               \\
\textbf{CUADPostTerminationServicesLegalBench}                                                                      & 60.02           & 57.43            & 55.94            & 60.64            & 59.78          & 58.91                & 57.55            & 57.92          & \textbf{61.14} & \underline{60.89}       & 59.28               \\
\textbf{CUADTerminationForConvenienceLegalBench}                                                                    & 80.93           & 79.07            & 79.53            & 83.26            & 67.91          & 77.44                & 80.70            & 77.21          & \textbf{84.65} & 84.42             & \textbf{84.65}      \\
\textbf{CzechSoMeSentiment}                                                                                         & 45.75           & 44.34            & 46.57            & 46.01            & 47.58          & \underline{47.62}          & 43.90            & \textbf{47.93} & 47.35          & 47.50             & 47.39               \\
\textbf{GujaratiNews}                                                                                               & 40.19           & 40.40            & 40.30            & 39.83            & \underline{40.55}    & \textbf{41.18}       & 39.07            & 39.77          & 40.08          & 39.27             & 38.92               \\
\textbf{HinDialect}                                                                                                 & 35.92           & 33.35            & 37.50            & \underline{38.67}      & \textbf{42.60} & 35.75                & 31.84            & 38.09          & 35.58          & 34.82             & 34.80               \\
\textbf{IndonesianIdClickbait}                                                                                      & 54.26           & 54.39            & 54.09            & 54.56            & \textbf{57.57} & 54.15                & 53.44            & \underline{55.73}    & 54.90          & 54.92             & 55.39               \\
\textbf{InternationalCitizenshipQuestionsLegalBench}                                                                & 57.47           & 57.32            & 56.59            & \textbf{62.01}   & 53.96          & 56.74                & 54.54            & 54.93          & 55.96          & 56.40             & \underline{57.96}         \\
\textbf{KLUE-TC}                                                                                                    & 21.16           & 20.39            & 21.88            & \underline{22.37}      & \textbf{23.34} & 22.06                & 21.41            & 22.13          & 21.27          & 21.40             & 21.86               \\
\textbf{Language}                                                                                                   & 92.56           & 91.40            & 93.83            & \underline{95.04}      & \textbf{96.05} & 92.80                & 93.13            & 93.22          & 92.23          & 92.42             & 93.28               \\
\textbf{LearnedHandsDivorceLegalBench}                                                                              & 76.00           & 80.67            & 75.33            & 69.33            & 64.67          & 80.67                & 83.33            & 82.00          & \textbf{85.33} & \underline{84.67}       & 84.00               \\
\textbf{LearnedHandsDomesticViolenceLegalBench}                                                                     & 78.16           & 73.56            & 78.74            & 70.69            & 72.41          & 75.86                & 75.29            & 74.71          & \textbf{81.03} & \underline{79.89}       & 77.01               \\
\textbf{LearnedHandsFamilyLegalBench}                                                                               & 70.75           & 72.41            & 68.65            & 71.48            & 64.99          & 68.85                & 71.53            & 76.95          & \underline{79.25}    & \textbf{79.98}    & 78.08               \\
\textbf{LearnedHandsHousingLegalBench}                                                                              & \textbf{74.76}  & \underline{73.34}      & 70.70            & 60.30            & 64.21          & 70.12                & 70.61            & 71.88          & 68.41          & 67.38             & 68.95               \\
\textbf{MacedonianTweetSentiment}                                                                                   & 35.77           & 35.66            & 37.44            & 36.77            & 37.95          & 37.12                & 36.50            & 37.85          & 37.36          & \textbf{38.01}    & \underline{37.98}         \\
\textbf{MarathiNews}                                                                                                & 36.24           & 36.68            & 37.33            & 37.47            & 37.52          & \textbf{38.23}       & 35.74            & 37.04          & 37.30          & 37.04             & \underline{37.63}         \\
\textbf{MassiveIntent}                                                                                              & 33.57           & 26.77            & 29.47            & 30.92            & 29.74          & 28.61                & \underline{33.57}      & 16.96          & \textbf{37.38} & 29.46             & 29.37               \\
\textbf{MassiveScenario}                                                                                            & \underline{35.86}     & 28.34            & 31.02            & 34.90            & 31.49          & 30.82                & 34.94            & 20.84          & \textbf{37.50} & 30.55             & 30.42               \\
\textbf{NorwegianParliament}                                                                                        & 52.46           & 52.60            & 52.25            & 52.83            & 51.33          & \textbf{53.24}       & \underline{52.89}      & 52.35          & 52.88          & 52.69             & 52.64               \\
\textbf{NYSJudicialEthicsLegalBench}                                                                                & 47.95           & 47.60            & 45.55            & 48.97            & 49.66          & 44.86                & 47.95            & \textbf{50.68} & \textbf{50.68} & 49.66             & 48.29               \\
\textbf{OPP115DataSecurityLegalBench}                                                                               & 71.21           & 71.96            & 70.91            & 73.16            & 58.17          & 69.94                & \underline{75.19}      & 73.54          & 74.51          & \textbf{75.64}    & 74.06               \\
\textbf{OPP115DoNotTrackLegalBench}                                                                                 & 81.82           & 86.36            & 78.18            & \underline{90.91}      & 80.91          & 80.00                & 81.82            & \underline{90.91}    & \textbf{91.82} & \underline{90.91}       & 87.27               \\
\textbf{\begin{tabular}[c]{@{}l@{}}OPP115InternationalAndSpecific\end{tabular}}               & 73.98           & \textbf{80.51}   & 78.88            & \underline{79.08}      & 76.73          & 78.27                & 77.04            & 76.33          & 74.18          & 73.37             & 75.00               \\
\textbf{OPP115PolicyChangeLegalBench}                                                                               & 87.24           & 87.70            & 86.54            & 88.40            & 83.06          & 88.17                & 84.45            & 86.77          & \textbf{89.79} & \textbf{89.79}    & \textbf{89.79}      \\
\textbf{OPP115ThirdPartySharingCollectionLegalBench}                                                                & 65.85           & 65.16            & 65.66            & 64.72            & 60.06          & 62.70                & \textbf{66.48}   & 64.78          & \underline{66.23}    & 64.65             & 65.79               \\
\textbf{OPP115UserChoiceControlLegalBench}                                                                          & 72.77           & 70.18            & 73.35            & 72.96            & \textbf{74.00} & 73.42                & \underline{73.67}      & 73.03          & 72.64          & 72.90             & 73.61               \\
\textbf{OralArgumentQuestionPurposeLegalBench}                                                                      & 22.44           & 21.79            & 21.47            & 19.87            & 24.04          & 23.08                & 21.79            & \textbf{25.32} & 24.68          & \underline{25.00}       & 23.08               \\
\textbf{PolEmo2}                                                                                                    & 34.27           & 32.67            & \textbf{37.35}   & 35.47            & \underline{36.84}    & 34.68                & 34.23            & 33.48          & 33.14          & 31.54             & 33.00               \\
\textbf{PunjabiNews}                                                                                                & 65.92           & 65.16            & 64.84            & 64.27            & \textbf{67.77} & 62.99                & 63.57            & 65.86          & 63.76          & \underline{66.88}       & 66.62               \\
\textbf{Scala}                                                                                                      & 50.14           & \underline{50.30}      & 49.98            & 50.21            & 50.15          & 50.26                & \textbf{50.37}   & 50.28          & 50.27          & 50.15             & 50.06               \\
\textbf{SCDBPTrainingLegalBench}                                                                                    & 62.80           & 59.37            & 59.37            & 56.99            & 51.72          & 55.94                & \underline{63.32}      & 62.27          & \textbf{64.38} & 61.74             & 61.74               \\
\textbf{SentimentAnalysisHindi}                                                                                     & 38.84           & 39.73            & \textbf{40.60}   & 39.11            & 39.47          & 39.65                & \underline{40.10}      & 38.82          & 38.70          & 38.90             & 39.16               \\
\textbf{SinhalaNews}                                                                                                & 35.97           & 35.80            & 35.46            & \underline{37.22}      & \textbf{39.90} & 35.14                & 34.82            & 34.15          & 33.72          & 33.12             & 34.35               \\
\textbf{SiswatiNews}                                                                                                & 71.25           & 71.63            & 73.13            & \textbf{74.50}   & 73.25          & 73.38                & 72.25            & \underline{73.50}    & 71.25          & 71.88             & 71.88               \\
\textbf{SlovakMovieReviewSentiment}                                                                                 & 52.71           & \textbf{53.85}   & 53.07            & 53.45            & 51.18          & 52.78                & 53.19            & \underline{53.65}    & 53.43          & 53.21             & 53.34               \\
\textbf{TamilNews}                                                                                                  & 18.25           & 18.21            & 18.25            & \underline{18.97}      & \textbf{19.27} & 18.50                & 17.79            & 18.36          & 18.05          & 17.87             & 17.76               \\
\textbf{TNews}                                                                                                      & 16.14           & 16.01            & 16.25            & \textbf{16.76}   & 16.56          & 16.45                & \underline{16.56}      & 15.28          & 15.19          & 15.43             & 16.02               \\
\textbf{TweetEmotion}                                                                                               & 26.47           & 26.19            & 28.48            & 28.32            & \textbf{29.40} & \underline{28.56}          & 26.08            & 26.72          & 27.23          & 27.15             & 27.55               \\ \toprule[1pt]
\textbf{Avg. Acc}                                                                                                   & 55.37           & 55.49            & 55.07            & 55.42            & 54.11          & 55.22                & 55.23            & 55.22          & \textbf{56.67} & \underline{56.11}       & 56.03              \\ \toprule[2pt]
\end{tabular}

\label{classifications}
\end{center}
\vspace{-2.5em}
\end{table*}

\subsection{Tasks}

Following the existing work, we first evaluate the model on seven English STS tasks(STS12\cite{sts12}, STS13\cite{13}, STS14\cite{14}, STS15\cite{15}, STS16\cite{16}, STS-B\cite{stsb}, SICKR\cite{sickr}) with the SentEval\cite{senteval} package; to assess the model's performance on STS tasks more comprehensively, we conduct experiments on the English subtasks of three multilingual STS tasks(STS17\cite{sts17}, STS22.V2\cite{sts22}, STSBenchmarkMultilingual\footnote{\url{https://github.com/PhilipMay/stsb-multi-mt}}) with the MTEB\cite{mteb} package.
In addition, we conduct a wide range of sentence embedding related 0-shot downstream tasks through the MTEB package, specifically including the multilingual or cross-language semantic text similarity computation tasks STS22.V2\cite{sts22}\footnote{STS22.v2 contains 19 subtasks, which is updated on STS22 by removing pairs where one of the entries contains empty sentences.}. Considering the limitation of computational resources, we randomly selected 45 text classification tasks, 45 text retrieval tasks, 15 bi-textmining tasks and the currently available data set of 14 text re-ranking tasks, totaling more than 130 subtasks, to demonstrate the robustness of JTCSE.

\subsection{Experimental Results}
We report the performance of JTCSE and baselines on the seven STS tasks in Table \ref{bigresult}, and overall, JTCSE-BERT and JTCSE-RoBERTa outperform the other work. Since JTCSE is a twin-tower structure, for a fair comparison with the single-tower model, we follow EDFSE and distill the knowledge of JTCSE through MSE loss to a naive BERT or RoBERTa denoted as JTCSE D. JTCSE D also outperforms the other single-tower baselines on average on the 7 STS tasks. Compared to the multi-tower model EDFSE, JTCSE-BERT has only one-third of the inference overhead of EDFSE-BERT but outperforms EDFSE-BERT on the STS tasks, which proves the effectiveness of the proposed modules constraint loss and cross-attention.

In addition, since RankCSE\cite{rankcse} is not officially open-sourced\footnote{All downstream experiments on RankCSE are reproduced from third-party open source code \url{https://github.com/perceptiveshawty/RankCSE}.}. RankCSE uses the same type of SimCSE-base\cite{gao2021simcse}, SimCSE-large, and DiffCSE-base\cite{diffcse} in constructing the teacher ensemble model, which relies on pre-existing work of the same type rather than having a BERT-base or RoBERTa-base autonomously trained to obtain the that directly comparing RankCSE with other work of the same type may lead to fairness discussions. Therefore, in order to make a fair comparison with RankCSE, we compare the performance on the 7 STS task by ensemble learning of JTCSE and RankCSE with the same unsupervised checkpoint InfoCSE\cite{infocse}, denoted as JTCSE-UC and RankCSE-UC, respectively. In addition, we distill JTCSE-UC and RankCSE-UC to a single naive encoder to obtain JTCSE-UC D and RankCSE-UC D, respectively; the experimental results show that JTCSE outperforms RankCSE on both -UC and -UC D.

In order to broadly evaluate the zero-shot performance of JTCSE and baselines on other natural language processing tasks, we conducted over 130 zero-shot tasks based on the MTEB evaluation package. We report the experiment results of text classification, text re-ranking, bi-textmining and multilingual semantic textual similarity in Table \ref{classifications}, Table \ref{reranking}, Table \ref{bitextmining}, and Table \ref{sts22}, in which we uniformly use MTEB's default “Main result” as the evaluation metrics; For the text retrieval task, due to the large number of metrics, in order to evaluate the performance of each model on the retrieval task more comprehensively, we adopt the following metrics, MAP@1/5/10, MRR@1/5/10, NDCG@1/5/10, PRECISION@1/5/10, and RECALL@1/5/10, and report the results\footnote{Compared to TNCSE's selection of MAP@10 as the evaluation metrics on 30 text retrieval tasks and JTCSE's selection of 5 categories with a total of 15 evaluation metrics to evaluate the model performance on 45 text retrieval tasks, JTCSE accomplished a more comprehensive evaluation and performed better on a broader range of tasks.} in Table \ref{retrieval}. On these four types of tasks, JTCSE and derived models are the best overall.

\subsection{Significance Test}

We use the official open-source code, specify random seeds from 1 to 5, and report the average results for 7 STS. The BERT-like single encoder is susceptible to random seeds; however, our proposed framework with a twin-encoder structure performs stably, and the average result is close to result in Table \ref{bigresult}. We report the significance test results in Fig. \ref{signification}.

\begin{table*}[h]
\setlength\tabcolsep{6pt}
\begin{center}
\caption{This table reports the results of the zero-shot evaluation of the 14 text reranking tasks in the MTEB package, which are all the reranking tasks for the currently available dataset. The optimal results on each task are \textbf{bolded}, and the sub-optimal results are \underline{underlined}. }
\vspace{-1em}
    \begin{tabular}{lccccccccccc}
\toprule[2pt]
\textbf{Tasks}                      & \textbf{Sim$\sim$} & \textbf{ESim$\sim$} & \textbf{Diff$\sim$} & \textbf{Info$\sim$} & \textbf{SN$\sim$} & \textbf{Whiten$\sim$} & \textbf{Rank$\sim$} & \textbf{TN$\sim$} & \textbf{JT$\sim$} & \textbf{JT$\sim$ D} & \textbf{JT$\sim$ UCD} \\ \toprule[1pt]
\textbf{Alloprof}          & 36.67           & 37.81            & 32.07            & 38.52            & 28.20          & 32.04                & 35.68            & 36.25          & \textbf{39.71} & \underline{39.63}       & 38.49               \\
\textbf{AskUbuntuDupQuestions}      & 51.88           & 52.28            & 52.08            & 52.83            & 45.53          & 51.60                & \underline{53.76}      & 50.73          & 52.85          & 52.65             & \textbf{54.01}      \\
\textbf{CMedQAv2}         & 13.97           & 14.78            & \underline{15.26}      & \textbf{17.21}   & 11.69          & 15.06                & 14.47            & 14.79          & 14.58          & 14.78             & 15.14               \\
\textbf{ESCI}              & \textbf{80.58}  & 80.28            & 80.49            & 80.36            & 78.05          & 80.47                & \underline{80.57}      & 79.75          & 80.17          & 80.51             & 80.54               \\
\textbf{MindSmall}         & 28.68           & 28.86            & \underline{29.34}      & 29.18            & 26.14          & 28.10                & \textbf{29.45}   & 28.65          & 28.76          & 28.75             & 28.92               \\
\textbf{MMarco}            & 2.48            & 3.77             & 3.64             & \textbf{4.96}    & 2.70           & 4.02                 & 3.34             & 2.94           & 4.02           & 4.04              & \underline{4.20}          \\
\textbf{NamaaMrTydi}       & \underline{39.88}     & 37.00            & 34.29            & 26.69            & \textbf{41.05} & 33.48                & 28.62            & 26.33          & 31.42          & 31.38             & 31.89               \\
\textbf{RuBQ}              & \textbf{27.33}  & 24.04            & 24.80            & 23.39            & 20.43          & \underline{25.34}          & 22.28            & 18.05          & 23.25          & 23.25             & 23.92               \\
\textbf{SciDocsRR}                  & 67.87           & 70.48            & 70.37            & \textbf{71.29}   & 58.90          & 67.63                & 69.89            & 70.51          & 69.85          & 69.59             & \underline{71.23}         \\
\textbf{StackOverflowDupQuestions}  & 39.56           & 40.63            & 42.77            & \textbf{44.21}   & 31.07          & 42.63                & 41.18            & 39.93          & 41.75          & 41.75             & \underline{43.35}         \\
\textbf{Syntec}            & 45.65           & 49.60            & 40.28            & 48.99            & 37.39          & 42.25                & 47.51            & 43.86          & \textbf{52.56} & \underline{50.93}       & 50.85               \\
\textbf{T2}                & 55.20           & 55.87            & 56.27            & 56.71            & 52.10          & 56.16                & 55.59            & 55.32          & 56.78          & \textbf{57.34}    & \underline{56.87}         \\
\textbf{VoyageMMarco}      & 21.60           & 21.41            & 20.90            & \textbf{23.57}   & 16.50          & 21.52                & 21.09            & 20.46          & 22.07          & 21.78             & \underline{22.69}         \\
\textbf{WebLINXCandidates} & 7.58            & 9.24             & 7.99             & 9.03             & 6.15           & 7.79                 & \underline{9.64}       & 8.82           & 9.25           & 8.71              & \textbf{10.26}      \\ \toprule[1pt]
\textbf{Avg. MAP}                       & 37.07           & 37.58            & 36.47            & \underline{37.64}      & 32.56          & 36.29                & 36.65            & 35.46          & \underline{37.64}    & 37.51             & \textbf{38.03}      \\ \toprule[2pt]
\end{tabular}

\label{reranking}
\end{center}
\vspace{-2em}
\end{table*}

\begin{table*}
\begin{center}
\caption{This table reports the results of the zero-shot evaluation of the 15 text bi-textmining tasks in the MTEB package, which are all the clustering tasks for the currently available dataset. The optimal results on each task are \textbf{bolded}, and the sub-optimal results are \underline{underlined}. }
\vspace{-1em}
\label{bitextmining}
\setlength\tabcolsep{7.5pt}
\begin{tabular}{lccccccccccc}
\toprule[2pt]
\textbf{Tasks}           & \textbf{Sim$\sim$} & \textbf{ESim$\sim$} & \textbf{Diff$\sim$} & \textbf{Info$\sim$} & \textbf{SN$\sim$} & \textbf{Whiten$\sim$} & \textbf{Rank$\sim$} & \textbf{TN$\sim$} & \textbf{JT$\sim$} & \textbf{JT$\sim$\ D} & \textbf{JT$\sim$\ UCD} \\ \toprule[1pt]
\textbf{BUCC}            & 0.55            & 1.54             & 0.54             & 0.60             & 0.12           & 0.25                 & 0.62             & \textbf{2.58}  & \underline{2.38}     & 2.23              & 1.56                \\
\textbf{BUCC.v2}         & 3.40            & 4.98             & 3.33             & 4.27             & 1.52           & 2.78                 & 4.21             & \textbf{7.24}  & \underline{7.15}     & 7.13              & 6.70                \\
\textbf{DiaBla}          & 4.08            & 5.55             & 3.71             & 4.36             & 2.07           & 3.56                 & 3.80             & \textbf{6.95}  & \underline{6.61}     & 6.55              & 4.98                \\
\textbf{Flores}          & 4.82            & \underline{5.50}       & 4.18             & 3.75             & 2.74           & 3.44                 & 5.04             & \textbf{5.56}  & 5.49           & 5.36              & 4.92                \\
\textbf{IN22Conv}        & 1.12            & 1.12             & 1.11             & \textbf{1.42}    & 1.06           & 1.11                 & 1.16             & 1.16           & \underline{1.23}     & 1.20              & 1.23                \\
\textbf{IN22Gen}         & 2.35            & 2.75             & 2.50             & \textbf{3.97}    & 2.01           & 2.67                 & 2.95             & 2.78           & 2.98           & 2.89              & \underline{3.09}          \\
\textbf{LinceMT}         & 15.44           & 15.65            & 15.53            & 15.22            & 4.43           & 16.22                & 14.45            & 16.30          & \underline{16.98}    & 16.68             & \textbf{17.10}      \\
\textbf{NollySenti}      & 18.76           & 19.78            & 19.01            & 22.33            & 10.12          & 19.44                & 18.95            & \underline{22.35}    & \textbf{22.61} & 21.85             & 22.05               \\
\textbf{NorwegianCourts} & 87.46           & 87.82            & 88.04            & 90.42            & 83.77          & 88.73                & 85.82            & 90.75          & \textbf{90.99} & \underline{90.96}       & 90.67               \\
\textbf{NTREX}           & 8.70            & 9.85             & 7.82             & 6.81             & 5.08           & 6.98                 & 8.96             & \textbf{10.58} & \underline{10.48}    & 10.18             & 9.19                \\
\textbf{NusaTranslation} & 45.52           & 45.93            & 44.61            & \textbf{50.36}   & \underline{50.31}    & 48.33                & 44.13            & 48.85          & 49.33          & 46.60             & 47.14               \\
\textbf{Phinc}           & 33.15           & 34.80            & 40.41            & \textbf{43.80}   & 27.58          & 41.79                & 38.13            & 41.43          & 41.33          & 39.53             & \underline{42.40}         \\
\textbf{RomaTales}       & 2.34            & 2.43             & 3.27             & 3.17             & 3.83           & 3.21                 & 2.00             & \textbf{4.43}  & 3.51           & \underline{4.11}        & 3.75                \\
\textbf{Tatoeba}         & 3.25            & 3.56             & 3.27             & 3.61             & 1.74           & 3.21                 & 3.43             & \textbf{4.31}  & \underline{4.23}     & 4.09              & 4.00                \\
\textbf{TbilisiCityHall} & 0.71            & 0.95             & 0.56             & 1.22             & 0.03           & 0.59                 & \textbf{1.46}    & 1.17           & 1.30           & \underline{1.41}        & 1.30                \\ \toprule[1pt]
\textbf{Avg. F1}         & 15.44           & 16.15            & 15.86            & 17.02            & 13.10          & 16.15                & 15.67            & \underline{17.76}    & \textbf{17.77} & 17.39             & 17.34               \\ \toprule[2pt]
\end{tabular}
\end{center}
\vspace{-2em}
\end{table*}

\begin{table*}[h]
\begin{center}
\caption{This table reports the evaluation results of each model on the multilingual or cross-language task STS22.V2, which contains 18 sub-test sets, with zero-shot evaluations for all languages or cross-languages except en. The optimal results for each sub-test set are \textbf{bolded}, and the sub-optimal results are \underline{underlined}.}
\setlength\tabcolsep{4.5pt}
\begin{tabular}{lccccccccccc}
\toprule[2pt]

\textbf{Tasks}    & \textbf{SimCSE} & \textbf{ESimCSE} & \textbf{DiffCSE} & \textbf{InfoCSE} & \textbf{SNCSE} & \textbf{WhitenedCSE} & \textbf{RankCSE} & \textbf{TNCSE} & \textbf{JTCSE} & \textbf{JTCSE\ D} & \textbf{JTCSE\ UCD} \\ \hline
\textbf{ar}       & \textbf{38.33}  & 32.48            & 34.94            & 21.08            & 33.58          & 36.08                & \underline{38.16}      & 34.75          & 35.16          & 32.77             & 33.15               \\
\textbf{avg}      & 32.82           & 36.79            & 34.37            & 28.09            & 23.64          & 32.71                & 38.49            & \textbf{39.33} & \underline{39.21}    & 37.76             & 37.82               \\
\textbf{de}       & 24.70           & \underline{28.50}      & 24.47            & 18.02            & 2.58           & 24.99                & 24.70            & 22.05          & 27.86          & 28.36             & \textbf{28.99}      \\
\textbf{de-en}    & 13.13           & 29.80            & 33.63            & \underline{37.03}      & 20.73          & 30.33                & \textbf{37.52}   & 33.10          & 36.33          & 30.84             & 34.76               \\
\textbf{de-fr}    & 35.93           & 32.68            & \underline{38.29}      & 2.44             & 25.42          & 31.45                & 37.81            & 35.41          & 32.52          & \textbf{40.43}    & 35.13               \\
\textbf{de-pl}    & 18.82           & 12.78            & 11.30            & -26.67           & 7.08           & 9.58                 & 5.67             & \textbf{36.71} & 23.13          & \underline{26.02}       & 17.68               \\
\textbf{en}       & 59.74           & 61.33            & 61.84            & 55.51            & 54.77          & 60.83                & 62.46            & 61.45          & 62.79          & \textbf{63.59}    & \underline{63.06}         \\
\textbf{es}       & 49.23           & 52.14            & 55.03            & 49.06            & 39.98          & 55.16                & 59.91            & \underline{61.34}    & \textbf{63.54} & 57.28             & 57.75               \\
\textbf{es-en}    & 30.44           & 37.84            & 36.83            & 38.53            & 21.28          & 34.14                & \textbf{39.37}   & 25.96          & \underline{38.77}    & 35.18             & 38.00               \\
\textbf{es-it}    & 31.48           & 42.50            & 40.91            & \underline{44.44}      & 22.54          & 31.27                & 42.43            & \textbf{45.70} & 42.56          & 43.80             & 44.14               \\
\textbf{fr}       & 61.55           & 61.31            & 60.06            & 52.95            & 31.47          & 52.96                & 64.85            & \textbf{67.70} & 64.22          & \underline{66.91}       & 65.50               \\
\textbf{fr-pl}    & 39.44           & \textbf{50.71}   & -5.63            & 16.90            & 16.90          & 16.90                & 39.44            & \textbf{50.71} & 28.17          & 28.17             & 28.17               \\
\textbf{it}       & 54.67           & 59.89            & 57.61            & 52.94            & 27.64          & 53.46                & 60.43            & 61.60          & \underline{62.37}    & 62.05             & \textbf{62.68}      \\
\textbf{pl}       & 22.79           & 26.72            & 23.77            & 8.23             & 6.78           & 23.42                & \textbf{31.00}   & 29.67          & \underline{30.46}    & 26.24             & 26.05               \\
\textbf{pl-en}    & 15.44           & \underline{36.41}      & 30.43            & 29.48            & 28.67          & 22.82                & 34.44            & 29.15          & \textbf{37.36} & 27.74             & 30.62               \\
\textbf{ru}       & 15.71           & 17.87            & 24.03            & 6.77             & 14.03          & \underline{24.59}          & 21.70            & \textbf{26.26} & 23.03          & 20.29             & 20.34               \\
\textbf{tr}       & 28.09           & 31.56            & 29.18            & 24.27            & 16.92          & 28.33                & 30.35            & 30.05          & \textbf{33.37} & \underline{31.80}       & 31.59               \\
\textbf{zh}       & 46.42           & 37.76            & \underline{48.78}      & 47.06            & 40.12          & 40.45                & \textbf{50.65}   & 39.64          & 48.23          & 41.81             & 42.71               \\
\textbf{zh-en}    & 4.82            & 9.87             & 13.14            & \textbf{27.61}   & 15.06          & 11.94                & 12.02            & 16.71          & 15.90          & 16.38             & \underline{20.49}         \\ \toprule[1pt]
\textbf{Avg. Acc} & 32.82           & 36.79            & 34.37            & 28.09            & 23.64          & 32.71                & 38.49            & \textbf{39.33} & \underline{39.21}    & 37.76             & 37.82               \\ \hline
\toprule[2pt]
\end{tabular}
\label{sts22}
\end{center}
\vspace{-2em}
\end{table*}

\begin{table*}
\setlength\tabcolsep{3.5pt}
\caption{This table reports the results of the zero-shot evaluation of JTCSE and baseline on MTEB, with the number of tasks and metric indicated in parentheses.
Details of the results are reported in Appendix \textbf{II}.}
\begin{tabular}{lccccccccccc}
\toprule[2pt]
{\textbf{Metrics}}       & \textbf{SimCSE} & \textbf{ESimCSE} & \textbf{DiffCSE} & \textbf{InfoCSE} & \textbf{SNCSE} & \textbf{WhinenedCSE} & \textbf{RankCSE} & \textbf{TNCSE} & \textbf{JTCSE} & \textbf{JTCSE D} & \textbf{JTCSE UCD} \\ \toprule[1pt]
{\textbf{MAP@1}}         & 7.16            & 8.43             & 7.60             & 6.70             & 2.77           & 7.57                 & 6.80             & 7.82           & \textbf{8.73}  & \textbf{8.73}    & \underline{8.58}         \\
{\textbf{MAP@5}}         & 10.04           & 11.32            & 10.38            & 9.28             & 4.16           & 10.42                & 9.48             & 10.76          & \underline{12.01}    & \textbf{12.03}   & 11.96              \\
{\textbf{MAP@10}}        & 10.79           & 12.07            & 11.10            & 9.93             & 4.54           & 11.16                & 10.19            & 11.47          & \underline{12.75}    & \textbf{12.77}   & 12.73              \\
{\textbf{MRR@1}}         & 11.81           & 13.12            & 11.99            & 11.14            & 4.50           & 11.84                & 10.61            & 11.92          & \underline{13.59}    & \textbf{13.62}   & 13.46              \\
{\textbf{MRR@5}}         & 16.02           & 17.48            & 16.21            & 15.43            & 6.71           & 16.24                & 14.79            & 16.41          & 18.39          & \textbf{18.54}   & \underline{18.40}        \\
{\textbf{MRR@10}}        & 16.90           & 18.38            & 17.09            & 16.21            & 7.27           & 17.12                & 15.63            & 17.30          & \underline{19.30}    & \textbf{19.41}   & 19.27              \\
{\textbf{NDCG@1}}        & 11.62           & 12.83            & 11.76            & 10.94            & 4.42           & 11.66                & 10.50            & 11.68          & \underline{13.36}    & \textbf{13.38}   & 13.25              \\
{\textbf{NDCG@5}}        & 13.91           & 15.27            & 14.29            & 13.21            & 5.77           & 14.25                & 13.12            & 14.55          & \underline{16.32}    & \textbf{16.38}   & 16.32              \\
{\textbf{NDCG@10}}       & 15.13           & 16.56            & 15.44            & 14.23            & 6.51           & 15.42                & 14.30            & 15.72          & 17.45          & \underline{17.49}      & \textbf{17.52}     \\
{\textbf{PRECISION@1}}   & 11.82           & 13.12            & 11.99            & 11.18            & 4.49           & 11.85                & 10.64            & 11.93          & \underline{13.59}    & \textbf{13.63}   & 13.46              \\
{\textbf{PRECISION@5}}   & 6.14            & 6.54             & 6.34             & 6.14             & 2.52           & 6.28                 & 5.89             & 6.26           & 7.08           & \textbf{7.15}    & \underline{7.13}         \\
{\textbf{PRECISION@10}}  & 4.64            & 4.92             & 4.72             & 4.61             & 1.88           & 4.71                 & 4.47             & 4.60           & \underline{5.20}     & 5.19             & \textbf{5.24}      \\
{\textbf{RECALL@1}}      & 7.16            & 8.43             & 7.60             & 6.70             & 2.77           & 7.57                 & 6.80             & 7.82           & \textbf{8.73}  & \textbf{8.73}    & \underline{8.58}         \\
{\textbf{RECALL@5}}      & 14.81           & 16.24            & 14.96            & 13.52            & 6.61           & 15.02                & 13.94            & 15.64          & 17.30          & \underline{17.38}      & \textbf{17.49}     \\
{\textbf{RECALL@10}}     & 19.51           & 20.90            & 19.49            & 17.66            & 9.32           & 19.48                & 18.28            & 20.24          & 21.88          & \underline{21.97}      & \textbf{22.15}     \\ \toprule[1pt]
{\textbf{45 Avg.}} & 11.83           & 13.04            & 12.06            & 11.12            & 4.95           & 12.04                & 11.03            & 12.27          & \underline{13.71}    & \textbf{13.76}   & 13.70     \\  \toprule[2pt]
\end{tabular}

\label{retrieval}
\vspace{-1em}
\end{table*}



\section{Ablation Studies}

\subsection{Alignment and Uniformity}

\begin{figure*}[htbp]
    \centering
    \begin{minipage}{0.32\textwidth}
        \centering
        \includegraphics[width=\textwidth]{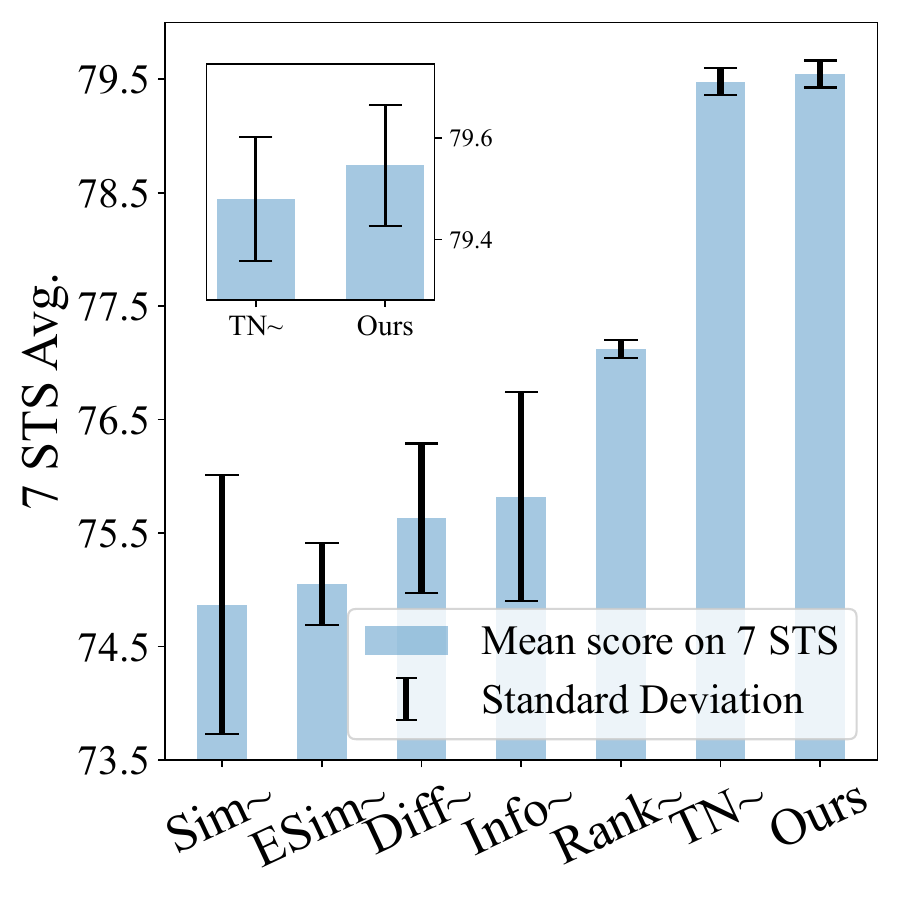}
        \caption{This figure reports the mean and variance of the results of the significance test experiments with random seed selection ranging from 1 to 5.}
        \label{signification}
    \end{minipage}
     \hspace{0.005\textwidth} 
    \begin{minipage}{0.33\textwidth}
        \centering
        \includegraphics[width=\textwidth]{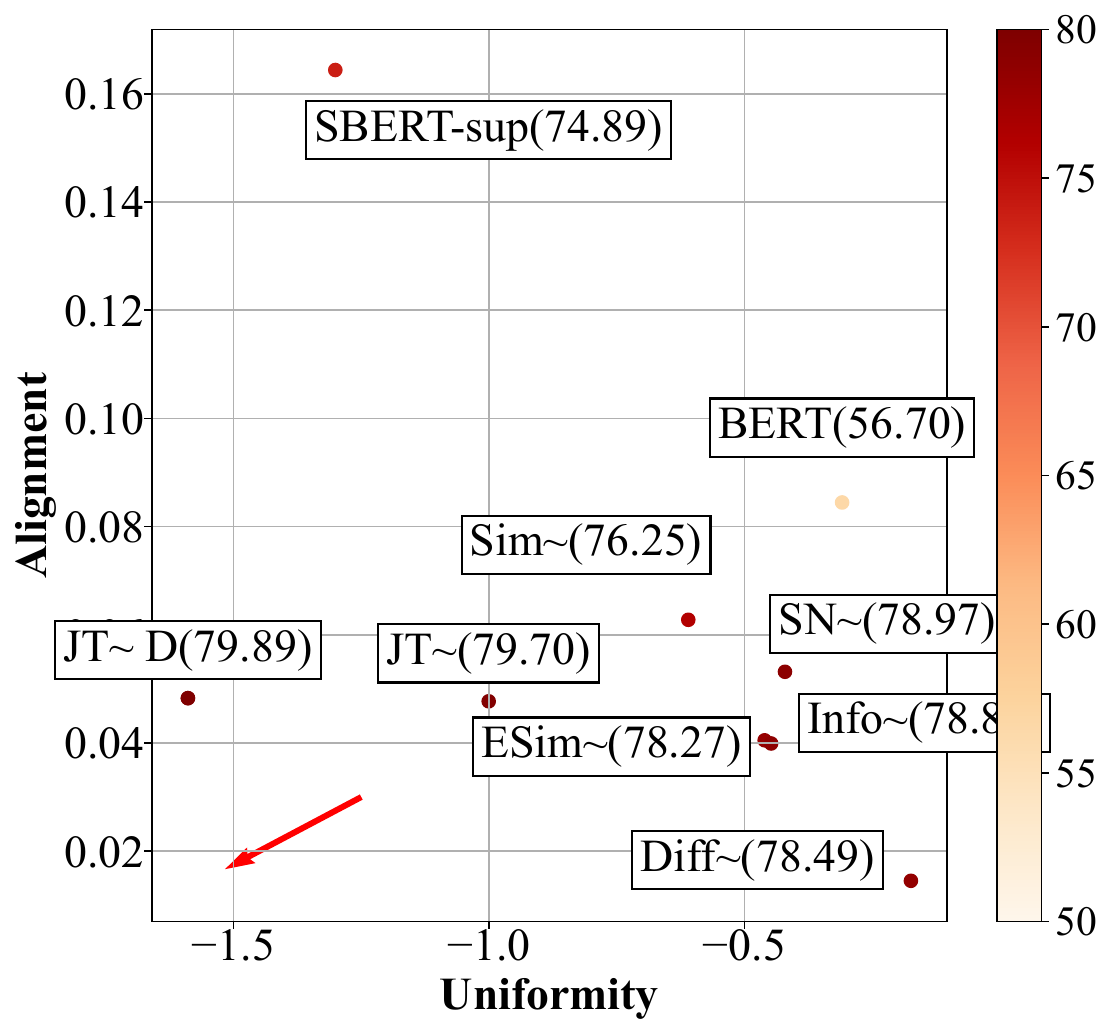}
        \caption{This figure reports uniformity and alignment metrics for JTCSE and baselines. Both of these metrics should be as small as possible and distributed as close to the bottom left as possible in this figure.}
        \label{uniformity}
    \end{minipage}
     \hspace{0.005\textwidth} 
    \begin{minipage}{0.315\textwidth}
        \centering
        \includegraphics[width=\textwidth]{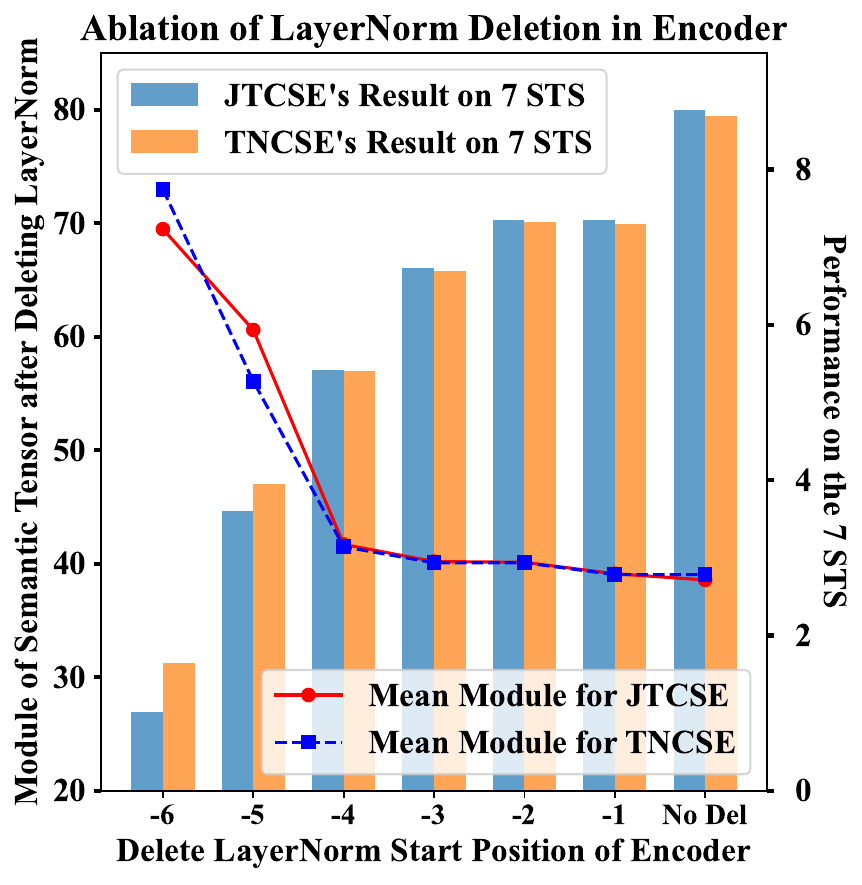}
        \caption{This figure reports the performance of JTCSE and TNCSE on 7 STS and the average modulus length of the output hidden states after removing some of the LayerNorms.}
        \label{disscuss_tmc}
    \end{minipage}
\end{figure*}

\cite{alignment-uniformity} proposes two critical evaluation metrics for evaluating the quality of embedding representations: Uniformity and Alignment. Uniformity means that the embedding representations of a mini-batch should be distributed as uniformly as possible on the unit hypersphere. Alignment means that two embedding representations that are positive samples of each other should be distributed as similarly as possible on the unit hypersphere. Both metrics should be as small as possible, denoted as Eq. \ref{D1} and Eq. \ref{D2}, respectively.
\begin{equation}
\begin{aligned}
l_{align} \triangleq  {\mathbf{E}}_{(x,x^{+}\sim p_{pos}  )}^{} \left \| f(x)-f(x^{+} ) \right \| ^{2} ,
\end{aligned}
\label{D1}
\end{equation}
where $\left \| \cdot  \right \| $ denotes the Euclidean paradigm, $f(x)$ denotes the embedding of the sample $x$ being projected. 
\begin{equation}
\begin{aligned}
l_{uniform} \triangleq \mathbf{log} \  {\mathbf{E}}_{i.i.d}^{} e^{-t\left \| f(x)-f(y) \right \| ^{2} } ,t>0,
\end{aligned}
\label{D2}
\end{equation}
where $x,y\sim p_{data}$ and $t$ is set to 2.
In Fig. \ref{uniformity}, we report the performance of the JTCSE and distillation model JTCSE D and the other baselines on these two metrics; the JTCSE series models outperform the other baselines overall.

\begin{table*}[]
\setlength\tabcolsep{2pt}
\caption{This table reports the experimental results, inference complexity and inference efficiency of different models in 7 STS.}
\begin{tabular}{ccccccccc|ccccc:ccc}
\hline
\textbf{Model}                                                                    & \textbf{Sim$\sim$} & \textbf{ESim$\sim$} & \textbf{Diff$\sim$} & \textbf{Info$\sim$} & \textbf{Rank$\sim$} & \textbf{EDFSE} & \textbf{TN$\sim$} & \textbf{JT$\sim$} & \textbf{Sim$\sim$} & \textbf{ESim$\sim$} & \textbf{Diff$\sim$} & \textbf{Info$\sim$} & \textbf{Rank$\sim$} & \textbf{EDFSE D} & \textbf{TN$\sim$ D} & \textbf{JT$\sim$ D} \\ \hline
                                                                                  & \multicolumn{8}{c|}{\textbf{Ensemble Model}}                                                                                            & \multicolumn{5}{c}{\textbf{Single-Tower Model}}                                        & \multicolumn{3}{c}{\textbf{Distilled Model}}           \\ \hline
\textbf{\begin{tabular}[c]{@{}c@{}}STS Acc\\ (7 Avg.)\end{tabular}}               & 77.97         & 78.51          & 77.89           & 77.05           & 78.22           & 79.04          & 79.58          & \textbf{79.70} & 76.25          & 78.27           & 78.49           & 78.85           & 76.04           & 79.27            & 79.77            & \textbf{79.89}   \\
\textbf{\begin{tabular}[c]{@{}c@{}}Inference Complexity\\ (GMAC)\end{tabular}}    & 10.90         & 10.90          & 10.90           & 10.90           & 10.90           & 32.70          & 10.90          & 10.90          & 5.40           & 5.40            & 5.40            & 5.40            & 5.40            & 5.40             & 5.40             & 5.40             \\
\textbf{\begin{tabular}[c]{@{}c@{}}Inference Efficiency\\ ($\eta$ )\end{tabular}} & 7.15          & 7.20           & 7.15            & 7.07            & 7.18            & 2.42           & 7.30           & \textbf{7.31}  & 14.12          & 14.49           & 14.54           & 14.60           & 14.08           & 14.68            & 14.77            & \textbf{14.79}   \\ \hline
\end{tabular}

\label{yita}
\vspace{-0.5em}
\end{table*}



\begin{figure*}
    \centering
    \includegraphics[width=1\linewidth]{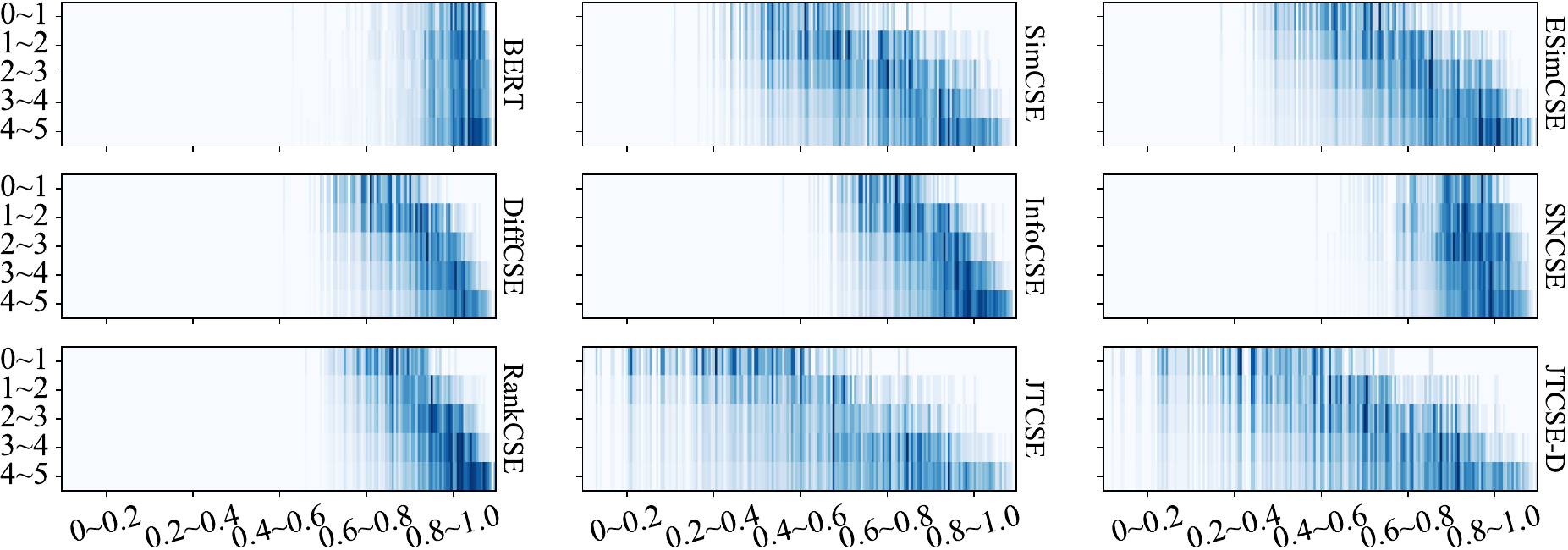}
    \caption{This figure reports the cosine similarity density distribution plots of different models on the STS-B dataset, where sentence pairs are uniformly divided into five groups, with the vertical coordinates of each subplot denoting the group and the horizontal coordinates denoting the model scoring. Each subplot should have an overall “sub-diagonal” distribution, indicating closer to the labeling distribution. We use the same code to report the performance of all models.}
    \label{cos}
    \vspace{-0.5em}
\end{figure*}

\subsection{Impact of Ensemble Learning and Analysis of Inference Efficiency}

Since JTCSE is a twin-tower structure, even though we have obtained its distillation to a single-tower model JTCSE D, to compare more fairly with other baselines, we used the same training set expansion method as JTCSE for each baseline\footnote{All of each baseline using the same RTT strategy, with the original training set being WIKI1M + unlabeled SICKR.} to train the sub-encoders and ensemble learning the obtained two sub-encoders, we report the evaluation results of the different ensemble models on the seven STS test sets in Table \ref{yita}, in addition, we report the evaluation results of JTCSE's direct ensemble learning of two subencoders before training. By comparison, the performance of the two sub-encoders is not optimal before being trained by JTCSE, and the performance is significantly improved after training, which indicates that ensemble learning does not play a central role in JTCSE becoming  SOTA on 7 STS tasks.

In addition, inference overhead is a key metric for practical applications of the models, and we report the inference overhead (GMAC\footnote{We use the Thop package to evaluate the inference overhead of the model.}) in Table \ref{yita}. In order to quantify the inference efficiency of each model, we define a simple metric for characterizing the model's performance per unit of inference overhead, defined as $\eta =\frac{\mathbf{Score} }{\mathbf{Cost} } $, where Score denotes the model's percentage of correctness on the seven STS test sets, and Cost denotes the model's inference complexity, reported in Table \ref{yita}. Among the multi-tower models, first notice that EDFSE adopts a naive multi-tower ensemble, which has a vast inference overhead and thus has a low performance per unit inference overhead; compared to other win-tower models, JTCSE has the highest performance per unit overhead as it performs the best on the 7 STS tasks, and among the single-tower models, as JTCSE D is a SOTA model, it has the $\eta $ highest.

\subsection{Impact of Unlabeled SICKR datasets}

We have found that when reproducing SimCSE, ESimCSE, and DiffCSE, we cannot reproduce the results reported in the paper using the official open-source code and default hyperparameters. For example, our reproduction of SimCSE is only about 74\%, which is far from the reported 76.25\%, so we can only improve the reproduction level by adding an unsupervised dataset, based on which we roughly improve the results of SimCSE to about 76\% in order to be fair enough to conduct the subsequent experiments.
In Table \ref{addingsickr}, we report the results we obtained by training the model with the default hyperparameters and the results by adding the SICKR dataset. In addition, we report the results of training JTCSE using the original Wiki1M to demonstrate that adding the unlabelled dataset does not significantly improve the model performance.
Meanwhile, we find that after the optimization of the cross-attention mechanism, the sensitivity of the SICKR dataset to the JTCSE is smaller than the TNCSE's. The gain of the SICKR dataset to the JTCSE is the smallest, which further demonstrates that the stability of the JTCSE is better.

\begin{table}[h]
\caption{The impact of adding unlabelled SICKR datasets on model training.}
\setlength\tabcolsep{4pt}
\begin{center}
\begin{tabular}{lccccccc}
\toprule[1pt]
\textbf{Datsets}          & \textbf{Sim$\sim $} & \textbf{ESim$\sim $} & \textbf{Diff$\sim $} & \textbf{Info$\sim $} & \textbf{Rank$\sim $} & \textbf{TN$\sim $} & \textbf{JT$\sim $}   \\ \hline
\textbf{Wiki 1M}          & 74.3         & 75.8          & 75.2          & 75.8          & 77.2          & 79.3        & \textbf{79.6} \\
\textbf{+SICKR} & 75.9         & 76.7          & 78.0          & 77.4          & 77.5          & 79.6        & \textbf{79.7} \\
\textbf{Gain$\downarrow $}             & 1.6          & 0.9           & 2.8           & 1.6           & 0.3           & 0.3         & \textbf{0.1}  \\ \toprule[1pt]
\end{tabular}

\label{addingsickr}
\end{center}
\vspace{-1em}
\end{table}

\subsection{The Ablation of Cross-Attention Structures}

In the cross-attention structure, we set CAELs at equal intervals, which means we set one CAEL for every pass through the same number of EncoderLayers.Thus, in JTCSE, the number of CAELs can be set to 1, 2, 3, 4, 6, and 12; if we do not set CAELs, the model degenerates to TNCSE. We report in Table \ref{Number of CAEL} the effect of setting different numbers of CAELs on the training of JTCSE.

When the number of CAELs is small, the features do not interact sufficiently across models to form an effective feature correction but instead may introduce error features, leading to attenuation of the model effect; when the number of CAELs is too large, the features are assimilated prematurely between sub-models, and it is unable to construct an adequate two-feature space for joint modeling, which then leads to the overfitting phenomenon similar to that of the single-tower model, and loses the significance of ensemble learning.

\begin{table}[hb!]
\caption{The effect of the number of CAELs on JTCSE training.}
\begin{center}
\setlength\tabcolsep{3pt}
\begin{tabular}{lccccccc}
\toprule[1pt]
\textbf{Number of CAEL} & \textbf{1} & \textbf{2} & \textbf{3} & \textbf{4} & \textbf{6}     & \textbf{12} & \textbf{w/o} \\ \hline
\textbf{7 STS Avg.}  & 79.23      & 78.74      & 79.57      & 79.37      & \textbf{79.70} & 79.55       & 79.58        \\ \toprule[1pt]
\end{tabular}

\label{Number of CAEL}
\end{center}
\end{table}

\subsection{Ablation on the Loss Function}

We have proposed the loss function of JTCSE in Model Structure Design, which consists of three parts: $L_{NCE}$, $L_{ICNCE}$, and $L_{ICTM}$.In this section, we analyze the gain from each part of the loss function and the reason for it, and we report the ablation results of this section in Table x.

\subsubsection{$L_{NCE}$ Only}
Since JTCSE employs a twin encoder structure, the sentence embedding is modeled through ensemble learning. If only unsupervised training of InfoNCE is performed for each encoder, which does not directly improve the ensemble learning, and result is improved insignificantly.

\subsubsection{$L_{ICNCE}$ Only}
Although ICNCE implements angle constraints for positive and negative samples between twin encoders and introduces cross-attention hidden state mutual supervision constraints, it does not substantially introduce a new training objective, so the effect improvement is still not obvious.

\subsubsection{$L_{ICTM}$ Only}
Since the training objective of the tensor modulus feature constraints proposed in Eq. \ref{eq11} is oriented to Pooler Output, but we use CLS Pooling in our inference, there is a margin between Pooler Output and CLS Pooling, so we cannot optimize CLS pooling directly, which in turn leads to an insignificant performance improvement.

\subsubsection{$L_{NCE}+L_{ICNCE}$}
Again, since no new training objective is introduced, strengthening the continuation training of each encoder and mutual supervision of twin encoders based only on the constraints of the tensor direction does not substantially improve the effectiveness.

\subsubsection{$L_{NCE}+L_{ICTM}$}
Under the joint effect of the continuation training of each encoder on the tensor direction constraint and the training goal of the mode length constraint, the model effect has been improved to a certain extent; however, without the introduction of the cross-attention can not effectively alleviate the phenomenon of attention sinking of the BERT-like model, so the model effect needs to be further improved.

\subsubsection{$L_{ICNCE}+L_{ICTM}$}
The model effect is significantly improved with the combined effect of cross-attention to alleviate attention sinking and tensor modulus constraints on the training objectives. Based on the ablation results obtained earlier, the role of the $L_{ICNCE}$, except for the cross-attention constraints, may overlap with that of the $L_{NCE}$ used to continue training. Thus, the model's performance in this setting is close to the final.

\begin{table}
\caption{This table combines each of the loss functions to explore the contribution of each to model training. \textbf{None} denotes a direct ensemble of two SimCSE-trained encoders. All experiments use CLS pooling method.}
\begin{center}
\begin{tabular}{lc}
\toprule[2pt]
\textbf{Loss Choice}         & \multicolumn{1}{l}{\textbf{7 STS Avg.}} \\ 
\toprule[0.75pt]
\textbf{None}(Twin Encoder Untrained)                  &78.27  \\
\textbf{$L_{NCE}$}                 & 78.50                                   \\
\textbf{$L_{ICNCE}$}               & 78.71                                   \\
\textbf{$L_{ICTM}$}                 & 78.40                                   \\
\textbf{$L_{NCE}+L_{ICNCE}$}           & 78.55                                   \\
\textbf{$L_{NCE}+L_{ICTM}$}             & 79.10                                   \\
\textbf{$L_{ICNCE}+L_{ICTM}$}           & 79.62                                   \\
\textbf{$L_{NCE}+L_{ICNCE}+L_{ICTM}$(Ours)} & \textbf{79.70}                          \\ 
\toprule[2pt]
\end{tabular}
\captionsetup{width=0.45\textwidth} 

\label{Loss Choice}
\end{center}
\vspace{-0.5em}
\end{table}

\begin{figure*}[h]
    \centering
    \includegraphics[width=1\linewidth]{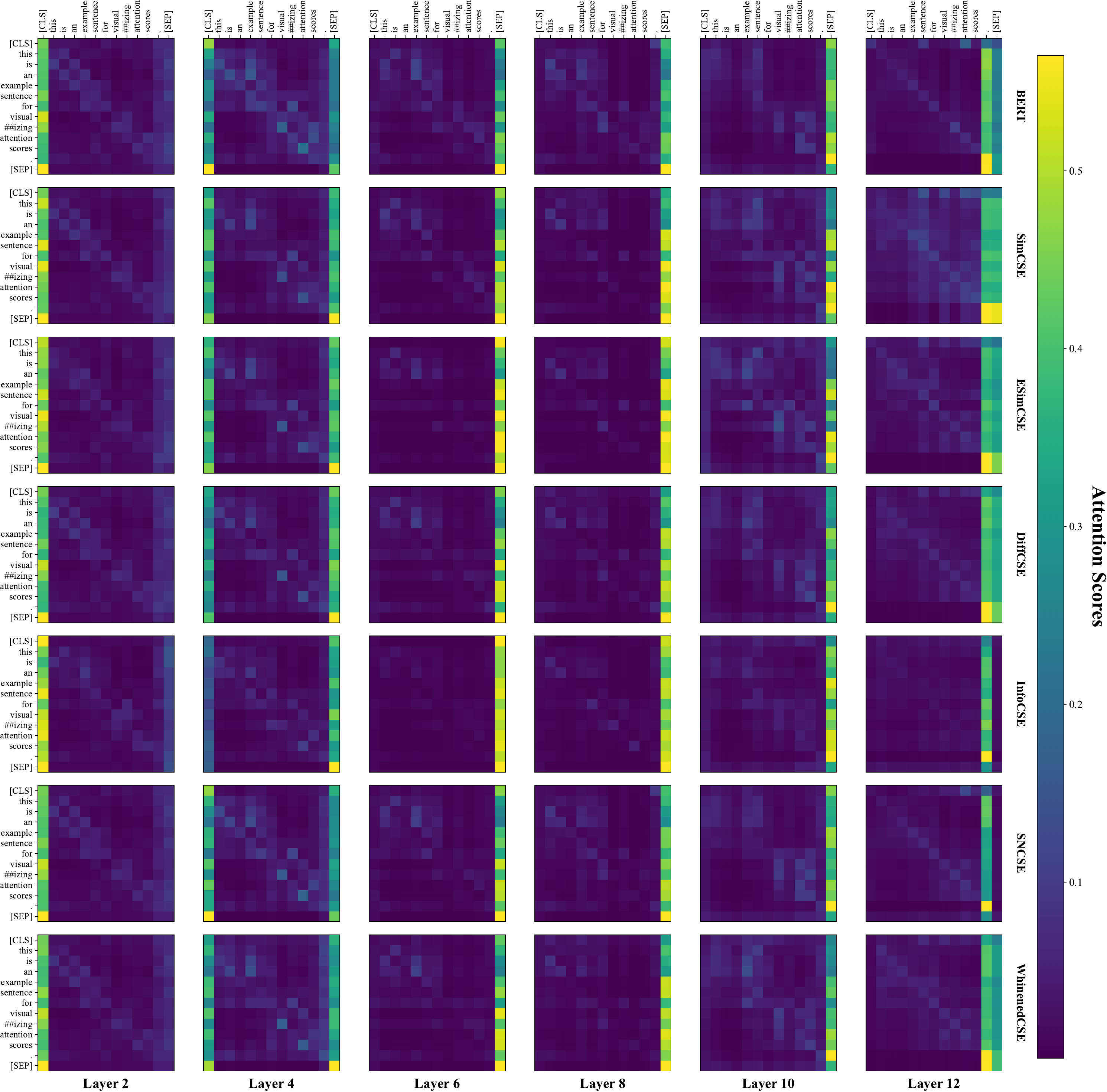}
    \caption{Visualization of attention scores across different layers and models (SimCSE, ESimCSE, DiffCSE, InfoCSE, SNCSE) for the input sentence: '\textbf{This is an example sentence for visualizing attention scores.}' Each subplot represents the average attention weights of a specific layer. As can be summarized from this figure, there is a clear attention sink for these representative models, which do not distribute the attention weights to the feature words in the sentence.}
    \label{attention_sink}
\end{figure*}

\section{Discussion}

\subsection{Reasonableness of the Tensor-Module Constrained Training Objectives' Design}
In our proposed training objective for tensor-constrained modulus length, we use the Pooler Outputs obtained after the last hidden state passes through the Pooler Layer for training. However, intuitively, it is more appropriate to use CLS pooling of the last hidden state for modulus length constraints consistent with the inference pooling approach, and in this section, we discuss why Pooler Outputs are used instead of the last hidden state.

The BERT-like models are all Encoder-Only structures and each EncoderLayer is structured like an encoder in a Transformer; specifically, each EncoderLayer contains a Self-Attention Mechanism module, a Forward Neural Network, and two LayerNorm.
LayerNorm is a normalization layer that normalizes the mean and variance of different hidden states. Due to LayerNorm, the forward-propagated hidden states will lose their modulus features. We observe SimCSE-BERT-base and find that no matter how much the semantics of the input sentences differ, the modulus of their output hidden state representations are always distributed in the range of 14$\sim$16, which makes the modulus constraints of the hidden state representations ineffective to be applied.

We further explore removing some of the LayerNorms during training to obtain the hidden state modulus features. By analyzing the output of 100 random sentences in Wiki100M, we find that removing the LayerNorm enhances the model's hidden state modulus features, but this operation significantly damages the model's performance.
Specifically, when training the TNCSE and JTCSE models, we gradually remove the LayerNorms in the penultimate 1 to 6 EncoderLayers and use the last hidden state after CLS pooling as the input for the modulus-constrained loss.
Fig. \ref{disscuss_tmc} shows the experimental results. As the number of removed LayerNorm layers increases, the model's performance on the seven STS tasks shows a systematic degradation. We hypothesize that this phenomenon stems from removing LayerNorm, destroying the key information learned in the pre-training stage of the BERT-base, and seriously degrading the model's semantic extraction capability. This finding suggests that although the presence of LayerNorm may limit the module feature of hidden states, it is crucial for maintaining the core capabilities of pre-trained language models.

The above discussion illustrates that the module's features of the hidden state cannot be obtained by removing the LayerNorm.We further investigate the structure of the BERT-like model and find a forward neural network named Pooler Layer after the last layer of EncoderLayer.To our knowledge, even though the Pooler Layer contains the BERT-like pre-training information, all BERT-like unsupervised sentence embedding models do not utilize the Pooler Layer, which wastes valuable pre-training knowledge.
We utilize the Pooler Layer and find that the Pooler Output obtained by processing the last hidden state through the Pooler Layer is characterized by modulus. Specifically, for different input sentences, the corresponding Pooler Output has an extensive distribution of modulus, which is almost unlimited, and it makes sense to perform modulus length constraints on this basis.

Therefore, based on the above discussion, we finally use the Pooler Output as the input of the tensor modulus constraint instead of the last hidden state.

\begin{table}[]
\caption{This table reports the effect of the pooling method on the model's performance on the 7 STS task. Avg is the average of all token hidden states in the last hidden state; Avg(FL) is the average of all token hidden states in the first and last layers; and Pooler is the output of the Pooler Layer.
All checkpoints are derived from official open source.}
\begin{center}
\setlength\tabcolsep{1pt}
\begin{tabular}{ccccccccccc}
\toprule[2pt]
Pooling Method & Sim           & ESim          & Diff          & Info          & whitened      & SN            & PCL           & Prompt        & JT            & JT-D          \\ \toprule[1pt]
CLS            & \textbf{76.3} & \textbf{78.3} & \textbf{78.5} & \textbf{78.9} & \textbf{78.8} & \textbf{79.0} & \textbf{78.4} & \textbf{41.1} & \textbf{79.7} & \textbf{79.9} \\
Avg            & 76.2          & 77.3          & 76.5          & 78.5          & 76.5          & 68.9          & 76.9          & 66.3          & 70.3          & 71.6          \\
Avg(FL)        & 75.5          & 75.5          & 72.4          & 74.5          & 72.4          & 69.7          & 74.1          & 66.6          & 78.4          & 78.4          \\
Pooler         & 75.3          & 67.2          & 78.2          & 78.5          & 78.1          & 50.7          & 78.0          & 22.7          & 70.3          & 71.6          \\ \toprule[2pt]
\end{tabular}

\label{pooling}
\end{center}
\vspace{-2em}
\end{table}

\subsection{Detailed Motivations for Cross-Attention Design}
By visualizing the attention weights, we observe that BERT-like models almost universally exhibit the attention-sinking phenomenon, as shown in the Fig.
 \ref{attention_sink}. The BERT-like model usually pays too much attention to the SEP token or the punctuation at the end of the sentence instead of the CLS token in the deep EncoderLayer.
However, all the unsupervised sentence embedding models use CLS pooling, and the CLS token is not focused on, which is detrimental to optimizing CLS pooling\footnote{Since the masked language model designed by BERT in the pre-training task does not mask the sequence's CLS token, the hidden state of the CLS token is considered to aggregate all the sequence's semantic information. We report the impact of the pooling approach in the Table \ref{pooling}}.
Moreover, several representative baselines demonstrate the phenomenon of attention sinking, which we hypothesize may be related to BERT's pre-training. Starting from the perspective of boosting the attention score may disturb BERT's pre-training information.
We can start from a different perspective by boosting the energy contained in the CLS token in a way to enhance its degree of being paid attention to, which in turn enriches the global semantic information aggregated by CLS pooling.

\begin{figure*}[ht]
    \centering
    \includegraphics[width=1\linewidth]{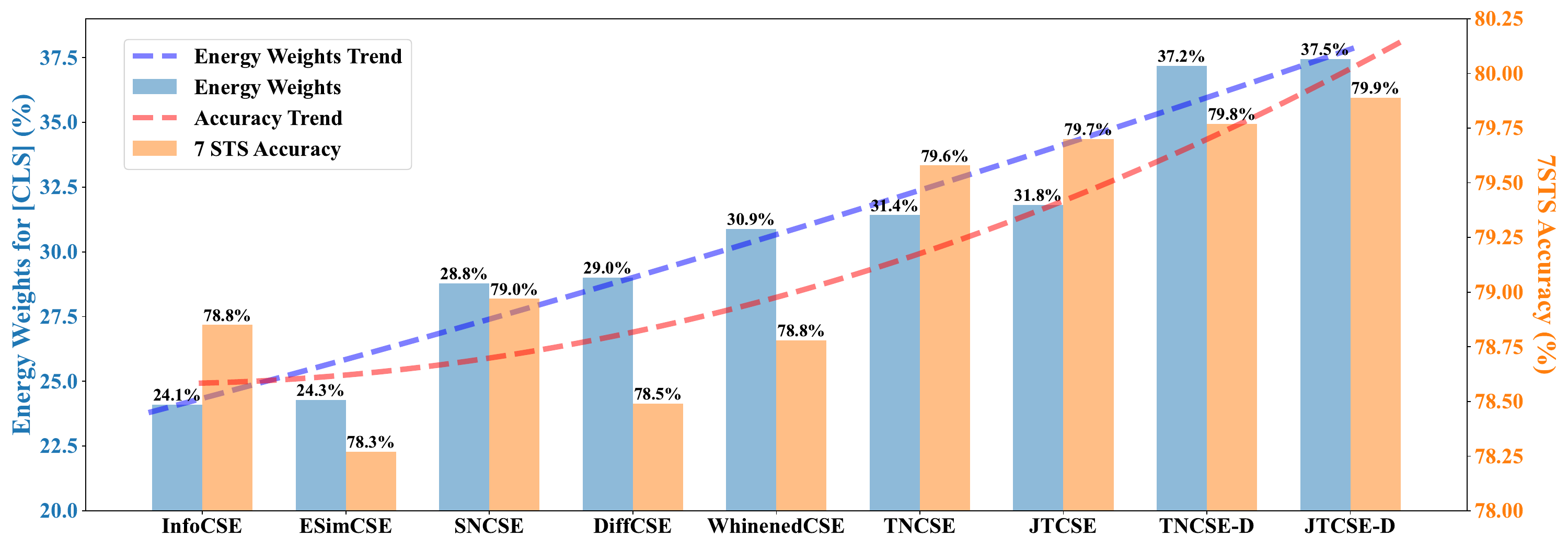}
    \caption{This figure illustrates that the model's performance on the 7STS task is approximately positively correlated with the energy weight occupied by the CLS token in the model's all-attention results.}
    \label{trend_line}
    \vspace{-1em}
\end{figure*}

After Query and Key compute the attention weight matrix, we note that the Value is weighted, and the weighted Value is defined as the context tensor. Naturally, we define the $E_{CLS}$ metric as Eq. \ref{CLS's energy weight}.
Intuitively, in Eq. \ref{CLS's energy weight}, the larger the $\left \| h_{cls}  \right \|_{2}$, the richer the CLS token aggregates semantic information, and the better CLS pooling effectiveness should be, which we demonstrate by exploring the relationship between the performance of some representative models on the 7 STS tasks and the $E_{CLS}$, as shown in Fig. \ref{trend_line}, which reports an almost positive correlation between the model's performance on the 7 STS tasks and the $E_{CLS}$.

To enhance the average CLS energy weight, we introduce a cross-attention structure within the twin encoder inspired by multimodal information fusion.
Cross-attention in twin encoders is a mechanism for information interaction between two different tensors, compared to traditional self-attention, which interacts with information based only on its own input sequence and may miss some important global information and lead to imperfect CLS pooling.
In contrast, cross-attention can achieve cross-encoder information sharing by using the attention weights of one encoder to weigh the Value tensor of another encoder.

Specifically, the Value of the external encoder provides additional contextual information that allows the CLS token of the current encoder to capture a richer representation of semantic features from another encoder. The reason for this is that cross-attention is computed across sub-encoders, and according to \cite{bertlookat}, the following possibility exists: the current encoder's attention weights may be more concerned with the syntactic features local to the sequence, whereas the external encoder's Value may contain richer global semantic features, and combines the two types of information when it is propagated forward to the next EncoderLayer of the current encoder, thus enhancing the the information density of the CLS.
In addition, this cross-attention design allows the model to dynamically adjust the information sources without changing the original attention distribution, which preserves the attention pattern of the current encoder without destroying the pre-training information inside the current encoder, and introduces new semantic complements through the Value of the external encoder. In particular, when the CLS token is required to serve as a global representation of the entire input sequence, the Value of the external encoder can provide a higher level of semantic support, making its hidden state more comprehensive and robust.

We clarify the proposed cross-attention designed to enhance $E_{CLS}$ and enrich the semantic information of CLS pooled aggregation by the above justifications.

\section{Conclusion}

In this work, we introduce the unsupervised sentence embedding representation framework JTCSE. In JTCSE, we first propose the training objective of tensor modulus constraints to improve the alignment between positive samples in unsupervised contrastive learning. Then, we introduce a cross-attention mechanism to optimize the quality of CLS Pooling to strengthen the model's attention to CLS tokens.
Through extensive evaluations, the results show that JTCSE is the current SOTA method for seven semantic textual similarity computation tasks and outperforms other models on hundreds of zero-shot evaluation tasks for natural language processing. In addition, we analyze the effects of important components in JTCSE through a series of ablation experiments. In future work, we will consider generalizing tensor mode length constraints and cross-attention mechanisms to multimodal learning tasks.
\bibliographystyle{IEEEtran}  
\bibliography{IEEEabrv,ieee_bib}


 

\begin{IEEEbiography}[{\includegraphics[width=1in,height=1.25in,clip,keepaspectratio]{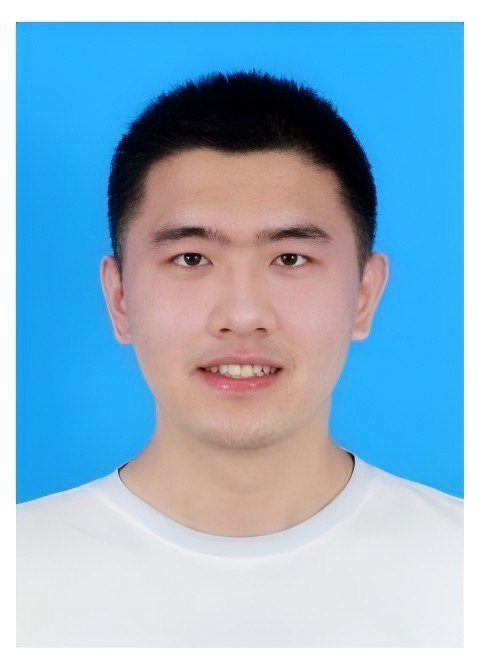}}]{Tianyu Zong} received his Bachelor's degree from the North China University of Technology in 2020 and his Master's in Electronic Information from the University of Chinese Academy of Sciences in 2023. He is pursuing a Ph.D. at the School of Computer Science and Technology, University of Chinese Academy of Sciences. His research focuses on natural language processing and multimodal information fusion. He has published two first-author papers at the AAAI conference and has been invited to give an oral presentation at AAAI 2025.

\end{IEEEbiography}
\vspace{-2em}
\begin{IEEEbiography}
[{\includegraphics[width=1in,height=1.25in,clip,keepaspectratio]{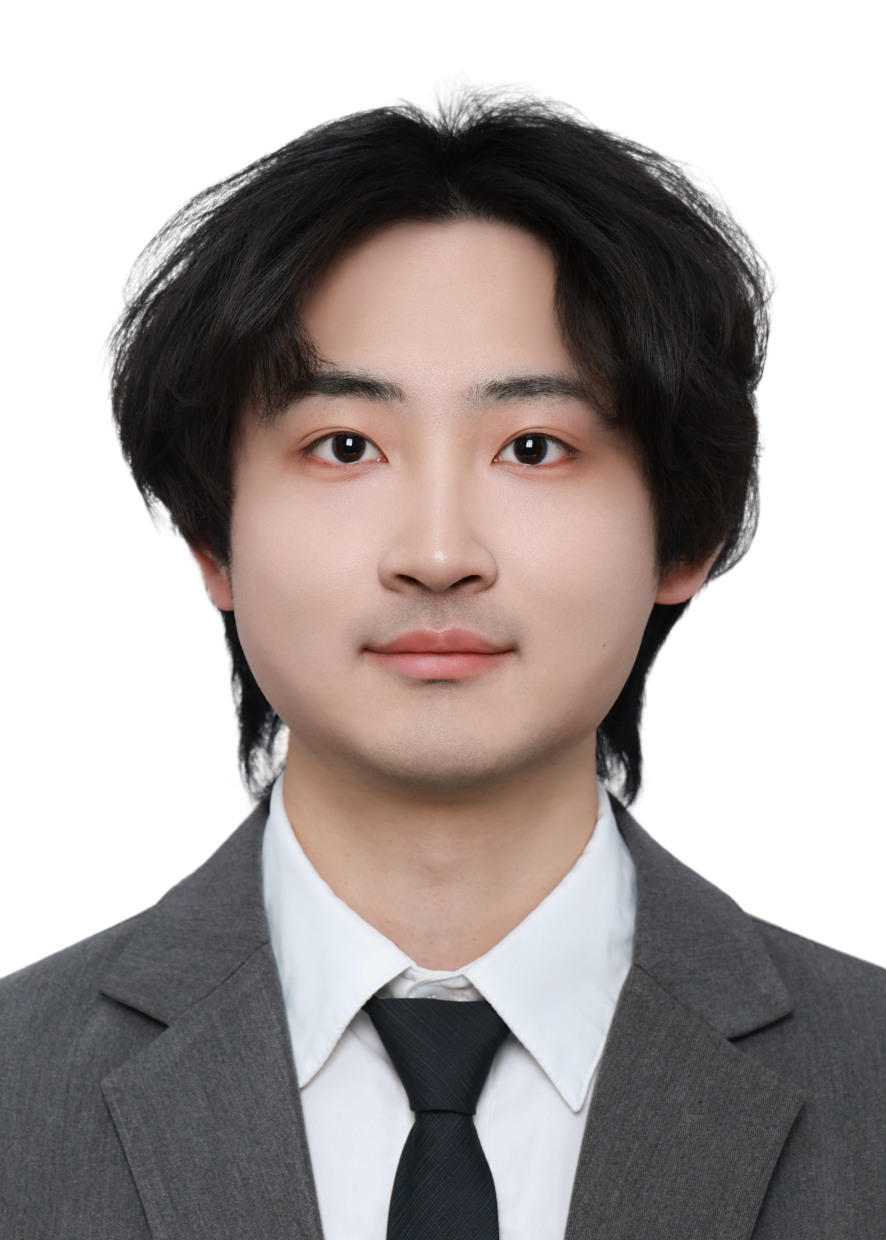}}]{Hongzhu Yi}
received his Bachelor's degree in Automation from Xidian University in 2023. He is currently pursuing a Ph.D degree at the Chinese Academy of Sciences. His research focuses on large models, with a particular emphasis on the understanding and generation of multimodal large models. He actively participates in various international competitions, including ICASSP MEIJU 2024, NeurIPS Edge LLMs Challenge 2024, CVPR Ego4D 2024, and CVPR Ego4D 2025.

\end{IEEEbiography}
\vspace{-1em}
\begin{IEEEbiography}
[{\includegraphics[width=1in,height=1.25in,clip,keepaspectratio]{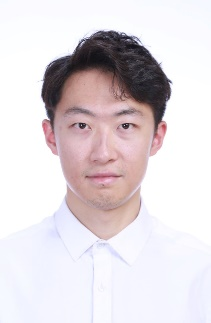}}]{Bingkang Shi} received his Bachelor's degree in Electronic and Information Engineering from Xidian University (XDU) in 2019. He is currently pursuing his PhD at the University of Chinese Academy of Sciences. His research focuses on Natural Language Processing (NLP), particularly model bias and natural language reasoning. His work has been published at ICASSP 2024 and EMNLP 2023. His honors include an Honorable Mention in the MCM/ICM competition and first prize in Xidian University's Spark Cup Science and Technology Competition.

\end{IEEEbiography}
\vspace{-1em}
\begin{IEEEbiography}
[{\includegraphics[width=1in,height=1.25in,clip,keepaspectratio]{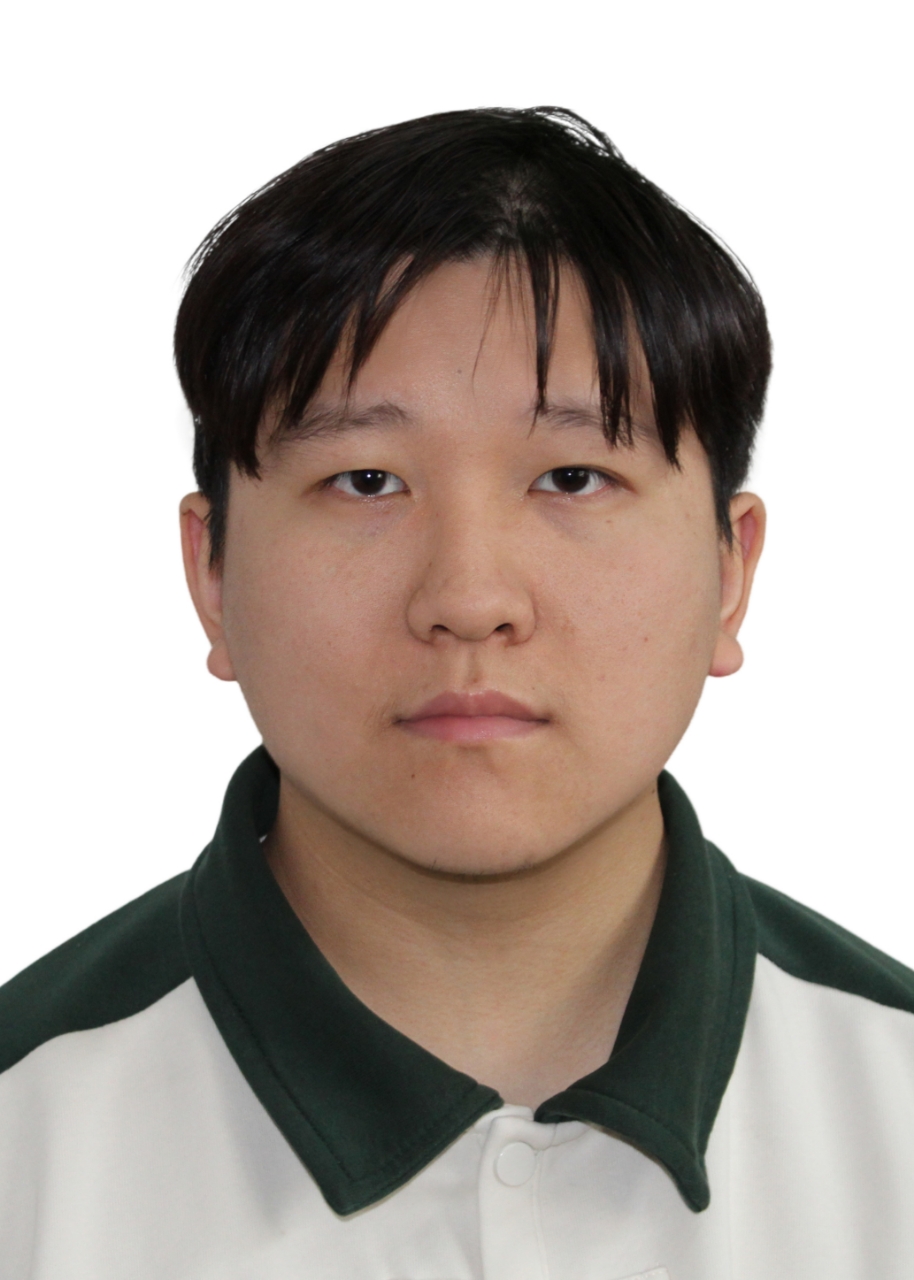}}]{Yuanxiang Wang} received the Bachelor's degree from the University of Chinese Academy of Sciences in 2023.He is currently working as an intern at the Cloud Computing and Intelligent Processing Laboratory (CCIP Lab) of UCAS.His research interests are focused on the field of multimodal large models, particularly on the fine-tuning and evaluation of these models.

\end{IEEEbiography}
\vspace{-1em}
\begin{IEEEbiography}
[{\includegraphics[width=1in,height=1.25in,clip,keepaspectratio]{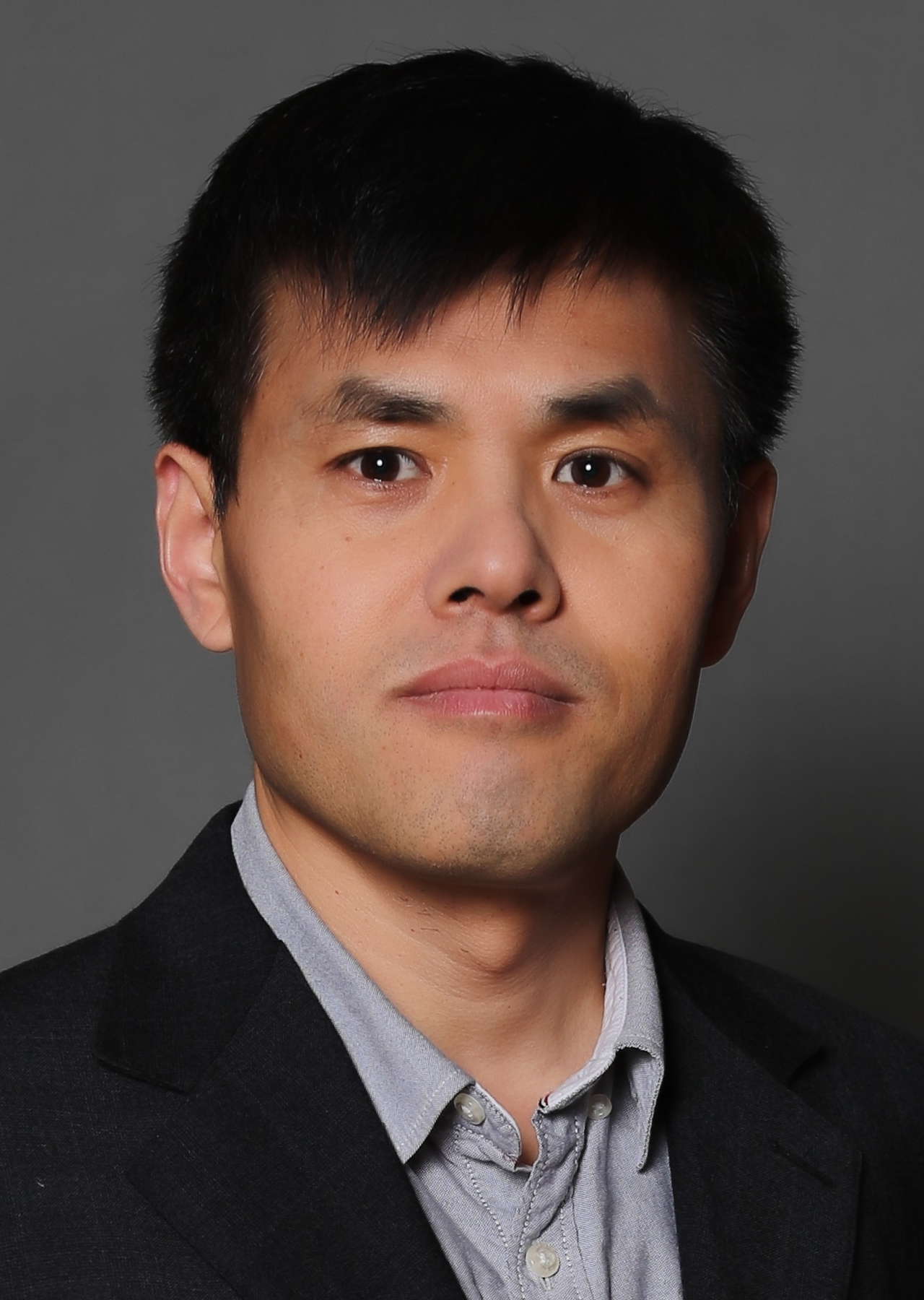}}]{Jungang Xu}
Jungang Xu is a full professor in School of Computer
Science and Technology, University of Chinese Academy
of Sciences. He received his Ph.D. degree in Computer
Applied Technology from Graduate University of Chinese
Academy of Sciences in 2003. His current research
interests include multi-modal intelligence, natural language processing and embodied intelligence.
He has published more than 30 papers in IEEE journals and conferences.
\end{IEEEbiography}



\vfill

\end{document}